\documentclass[10pt,twocolumn,letterpaper]{article}

\usepackage[pagenumbers]{cvpr} 

\usepackage{graphicx}
\usepackage{amsmath}
\usepackage{amssymb}
\usepackage{booktabs}
\usepackage[nolist]{acronym}

\usepackage{multirow}
\usepackage{arydshln}
\usepackage[accsupp]{axessibility}

\usepackage[pagebackref,breaklinks,colorlinks]{hyperref}

\usepackage[capitalize]{cleveref}
\crefname{section}{Sec.}{Secs.}
\Crefname{section}{Section}{Sections}
\Crefname{table}{Table}{Tables}
\crefname{table}{Tab.}{Tabs.}

\def\cvprPaperID{10744} 
\def\confName{CVPR}
\def\confYear{2023}

\begin{document}
\newacro{TTS}[TTS]{text-to-speech synthesis}
\newacro{SOTA}[SOTA]{State-of-the-art}
\newacro{ASR}[ASR]{Automatic Speech Recognition}
\newacro{GAN}[GAN]{Generative Adversarial Nets}
\newacro{HRI}[HRI]{Human Robot Interaction}
\newacro{TFG}[TFG]{Talking Face Generation}
\newacro{FGLES}[TalkLip net]{Face Generation under a Lip-reading Expert's Supervision}
\newacro{CNN}[CNN]{Convolutional Neural Network}
\newacro{HMM}[HMM]{hidden Markov models}
\newacro{ROI}[ROI]{Region of interest}
\newacro{WER}[WER]{Word Error Rate}
\newacro{MFCC}[MFCC]{Mel-Frequency Cepstral Coefficients}

\renewcommand{\thefootnote}{\fnsymbol{footnote}}
\let\origthanks\thanks
\renewcommand{\thanks}[1]{\renewcommand{\thefootnote}{$\dagger$}\origthanks{#1}\renewcommand{\thefootnote}{\arabic{footnote}}}

\title{Seeing What You Said: Talking Face Generation Guided\\ by a Lip Reading Expert}


\author{Jiadong Wang$^{1*}$,  Xinyuan Qian$^{2\dagger}$, Malu Zhang$^3$, Robby T. Tan$^1$, Haizhou Li$^{4,1}$\\
$^1$National University of Singapore, $^2$University of Science and Technology Beijing\\ $^3$University of Electronic Science and Technology of China \\
$^4$SRIBD, School of Data Science, The Chinese University of Hong Kong, Shenzhen, China\\ 
}



\maketitle

\begin{abstract}
    Talking face generation, also known as speech-to-lip generation, reconstructs  facial motions concerning lips given  coherent speech input. 
    The previous studies revealed the importance of lip-speech synchronization and visual quality.
    Despite much progress,
    they hardly focus on the content of lip movements i.e., the visual intelligibility of the spoken words, which is an important aspect of generation quality.
    To address the problem, we propose using a lip-reading expert to improve the intelligibility of the generated lip regions by penalizing the incorrect generation results.
    Moreover, to compensate for data scarcity, we train the lip-reading expert in an audio-visual self-supervised manner. With a lip-reading expert, we propose a novel contrastive learning to enhance lip-speech synchronization, and a transformer to encode audio synchronically with video, while considering global temporal dependency of audio.
    For evaluation, we propose a new strategy with two different lip-reading experts to measure intelligibility of the generated videos. 
    Rigorous experiments show that our proposal is superior to other \ac{SOTA} methods, such as Wav2Lip, in reading intelligibility i.e.,  over {\bf 38\%} \ac{WER} on LRS2 dataset and {\bf 27.8\%} accuracy on LRW dataset.
    We also achieve the \ac{SOTA} performance in lip-speech synchronization and comparable performances in visual quality. 
    \footnotetext{$^{*}$ jiadong.wang@u.nus.edu}
    \footnotetext{$^{\dagger}$ corresponding author (qianxy@ustb.edu.cn)}
\end{abstract}
\vspace{-6mm}

\section{Introduction}
\label{sec:intro}

\ac{TFG} aims at generating high-fidelity talking heads which are temporally synchronized with the input speech. It plays a significant role in many \ac{HRI} applications, such as film dubbing~\cite{kim2018deep}, video editing, face animation~\cite{song2018talking,gu2020flnet}, and communication with people who has hearing loss but master in lip-reading. Thanks to its various practical usage,
\ac{TFG} has also received an increasing attention in both industrial and research community over the past decades~\cite{lele2019hierarchical,jamaludin2019you}.


In \ac{TFG}, there are two major aspects of concerna: lip-speech synchronization and visual quality. While humans are sensitive to subtle abnormalities in asynchronized speech and facial motions~\cite{yu2020multimodal}, the mechanism of speech production highly relies on lip movements~\cite{ling2010analysis}. As a result, the main challenge of \ac{TFG} exists in temporal alignment between the input speech and synthesized video streams. 
One solution to this problem is to place an auxiliary embedding network at the end of the generator to analyze the audio-visual coherence. For example, \cite{prajwal2020lip} uses a pre-trained embedding network \cite{chung2017out} as the lip-sync discriminator, while \cite{zhu2018arbitrary} investigates an asymmetric mutual information estimator.
Another solution is to compute a sync loss between visual lip features of ground truth and generated video sequences~\cite{park2022}.
Other attempts use the encoder-decoder structure to facilitate \ac{TFG} by improving audio and visual representations, such as~\cite{zhou2019talking,zhou2021pose}, which disentangle the visual features to enhance audio representations into a shared latent space; or, \cite{mittal2020animating}, which disentangles audio factors, i.e., emotional and phonetic content, to remove sync-irrelevant features.



Apart from lip-speech synchronization, blurry or unrealistic visual quality also penalizes generation performance.
To preserve defining facial features of target persons, skip connections \cite{ronneberger2015u} are applied \cite{jamaludin2019you}. 
Others~\cite{zhou2019talking,chen2019hierarchical} employ \ac{GAN} to distinguish real and synthesized results by modelling the temporal dynamics of visual outputs. In this way, the resulting models can generate more plausible talking heads that can be qualitatively measured by subjective evaluations.

In addition, reading intelligibility should be indispensable, but it has not been emphasized. 
Reading intelligibility indicates how much text content can be interpreted from face videos by humans' lip reading ability, which is especially significant for hearing-impaired users. 
 However, image quality and lip-speech synchronization do not explicitly reflect reading intelligibility. Specifically, a well-qualified image may contain fine-grained lip-sync errors~\cite{prajwal2020lip} while precise synchronization may convey incorrect text contents. According to the McGurk effect\cite{mcgurk1976hearing}, when people listen and see an unpaired but synchronized sequence of speech and lip movements, they may recognize a phoneme from audio or video, or a fused artifact.


In this paper, we propose a TalkLip net to synthesize talking faces by focusing on reading intelligibility. 
Specifically, we employ a lip-reading expert which transcribes image sequences to text to penalize the generator for incorrectly generated face images. We replace some images of a face sequence with the generated ones, and feed it to the lip-reading expert during training to supervise the face generator.

However, lip reading is hard even for humans. In \cite{hassanat2011visual}, four people with equal gender distribution are invited to read lip movements. However, the average error rate is as high as 47\%. Therefore, a reliable lip-reading model relies on a great amount of data. We employ AV-Hubert\cite{shi2022learning}, a self-supervised method, which has yielded \ac{SOTA} performance in lip-reading, speech recognition, and audio-visual speech recognition. The encoders of lip-reading and speech recognition systems are highly synchronized since they are supervised by the same pseudo label during pre-training. 

Leveraging the lip-reading expert from the AV-Hubert, we propose a new method to enhance lip-speech synchronization. Particularly, we conduct contrastive learning between audio embeddings for face generation and visual context features from the lip-reading encoder. Besides, the AV-Hubert also provides a synchronized speech recognition system whose encoder considers long-term temporal dependency, we adopt this encoder to encode audio inputs, instead of encoders in \cite{prajwal2020lip, park2022, zhou2021pose} only rely on short-term temporal dependency (0.2s audio), or the single-modality (audio) pre-trained encoder in \cite{fan2022faceformer}. Our contributions are summarized as follows:
\vspace{-0.6mm}
\begin{itemize}
    \item We tackle the reading intelligibility problem of speech-driven talking face generation by leveraging a lip-reading expert.
   \vspace{-0.6mm}
    \item To enhance lip-speech synchronization, we propose a novel cross-modal contrastive learning strategy, assisted by a lip-reading expert.
    \vspace{-0.6mm}
    \item We employ a transformer encoder trained synchronically with the lip-reading expert to consider global temporal dependency across the entire audio utterance. 
    \item We propose a new strategy to evaluate reading intelligibility for \ac{TFG} and make the benchmark code publicly available\footnote{Code link: \textcolor{magenta}{\url{https://github.com/Sxjdwang/TalkLip}}}.  
    \vspace{-0.6mm}
    \item Extensive experiments demonstrate the feasibility of our proposal and its superiority over other prevailing methods in reading intelligibility (over 38\% \ac{WER} on LRS and 27.8\% accuracy on LRW). Additionally, our approach performs comparably to or better than other \ac{SOTA} methods in terms of visual quality and lip-speech synchronization.
\end{itemize}


\section{Related work}
\label{sec:related}

\subsection{Speech-driven talking face generation}
\ac{TFG} was first studied in the 1990s \cite{yehia1998quantitative} which aims at mapping acoustic features to time-aligned realistic facial motions. Traditional methods employ \ac{HMM}~\cite{bregler1997video}, such as \cite{brand1999voice} predicts facial motions through \ac{HMM} learnt from time-aligned audio-visual streams with the minimized entropy.
Recently, deep learning achieved great success which also brought benefits to the \ac{TFG} task.
Indeed, the majority of those \ac{TFG} works can be categorized into intermediate representation-based and reconstruction-based methods~\cite{park2022}, as elaborated next.

\begin{figure*}[!htb]
    \centering
    \includegraphics[width=\textwidth]{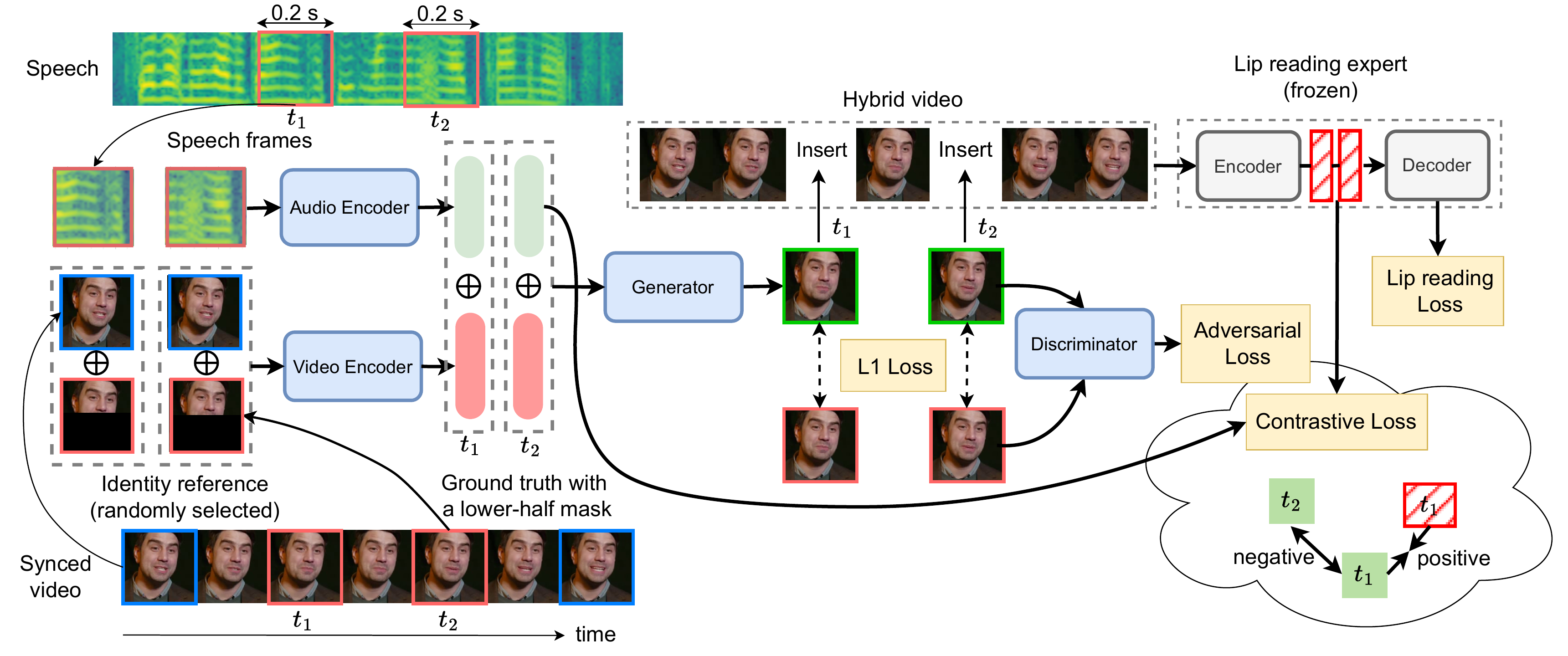}
    \caption{Block diagram of our proposed TalkLip net which produces talking faces given the coherent speech inputs, identity and pose inputs. The network consists of (1) an audio encoder for phoneme-level feature extraction; (2) a video encoder to extract the identity and pose information; (3) a \ac{GAN} conditioned on the real/synthesized face images and (4) a novel back-end lip-reading expert to penalize the lip-reading error. The whole network is optimized with the four-fold losses, which are marked as yellow rectangles ($\bigoplus$ denotes pairwise concatenation). 
    }
    \label{fig:arch}
    \vspace{-5mm}
\end{figure*}

Intermediate representation-based \ac{TFG} methods include two cascaded modules which learn intermediate facial representations i.e., landmarks or 3D meshes for face synthesis from the coherent speech input. For example, in \cite{chen2019hierarchical}, facial landmarks are predicted from input speeches which are then used to generate face videos conditioned on a reference image.
In \cite{zhou2020makelttalk}, landmarks dependent on speech content and speaker identity are disentangled to generate faces of unseen subjects. 
In \cite{song2022everybody}, 3D facial meshes are reconstructed from the extracted expression, geometry and pose parameters to facilitate \ac{TFG}.
However, these methods strongly depend on the quality of 3D face model construction, and are hard to be applied to arbitrary identities~\cite{zhou2019talking}. Thus has limited generalization capability.  Moreover, fine details i.e., teeth regions are difficult to be captured using 3D models~\cite{zhang2021flow} which may degrade the generation performance.

Reconstruction-based methods mostly adopt the encoder-decoder structure to generate talking face videos end-to-end. In \cite{prajwal2020lip}, a pre-trained discriminator conditioned on the lip-speech sync is employed which emphasizes its impact on the sync quality.  By disentangling speech- and identity-related factors from video, \cite{zhou2019talking} learns better visual representations through the proposed associative and adversarial training strategy.
Differently, \cite{mittal2020animating} improves the audio representations by disentangling phonetic and emotional content from speech to generate better \ac{TFG} results. Reconstruction-based methods are free of generating intermediate ground truth and contain fine details of mouth \cite{park2022}. 
Therefore, we rely on the reconstruction strategy to synthesize photo-realistic talking faces from speech directly.

To be noted, our concept of employing a lip-reading expert to penalize the incorrectly generated images can be easily extended to both categories while we select the reconstruction strategy to set an example.

\subsection{Contrastive learning}
Contrastive learning is a strategy to obtain more representative features and has widespread usage in various tasks, such as text-image retrieval~\cite{li2022clip}, image classification~\cite{wang2021contrastive} and acoustic event detection~\cite{wu2022wav2clip}.
Its core idea is minimizing the distance between samples of the same class while maximizing the ones from different classes.
To this end, one solution is by mapping all features (either homogeneous or heterogeneous) into a common representation space where \textit{positive} feature pairs are attracted and \textit{negative} pairs are repelled. 
Indeed, selecting {positive} and {negative} pairs is of particular significance, which guides the model to learn contrastive representations.
It also promotes self-supervised learning without the requirement of labeled data, such as SimCLR~\cite{chen2020simple} and SwAV~\cite{caron2020unsupervised}.

In talking face generation, contrastive learning can assist lip-speech synchronization. Concretely, PC-AVS \cite{zhou2019talking} uses audio features that are timely aligned with the visual lip feature as positive samples and adopt others as negative ones. However, they extract visual lip features only from single images without considering lip movement. Unlike PC-AVS, our contrastive learning focuses on the dynamic property of lip movement and extracts visual features from 5 frames.

\section{Proposed Methods}\label{sec:method}
We propose a TalkLip net, illustrated in Fig.~\ref{fig:arch}, to improve the reading intelligibility of the generated talking faces videos. 
Specifically, the TalkLip net takes images as the identity and pose references while taking the coherent speech as the lip movement reference. Given synthesized videos from the generator, the lip-reading expert is employed to penalize inaccurate lip movements via a lip-reading loss. This will backpropagate gradients to the generator, the audio and video encoders, as described in \cref{subsec:lip_loss}. Furthermore, assisted by the lip-reading expert, we perform contrastive learning between visual context features and audio embeddings to improve lip-speech synchronization, as depicted in \cref{subsec:contrastive_loss}. Finally, all losses
are integrated in~\cref{subsec:contrastive_loss} to facilitate \ac{TFG}.



\subsection{Overview of lip-reading expert}
\label{subsec:lipreading}
The goal of a lip-reading system is to transcribe a word sequence (or something like that) from the lip movement in a video. To tackle data scarcity, we adopt an audio-visual self-supervised method, AV-Hubert \cite{shi2022learning} to have a reliable lip-reading expert. 

The self-supervision is shown in \cref{fig:transformer}. The visual front-end consists of a 3D \ac{CNN} to capture local lip movement and a 2D ResNet-18 \cite{stafylakis2017combining}. The modality selector determines the training of audio-only, video-only or audio-visual speech recognition systems by masking features of no-involved modalities. All systems share the same transformer encoder and are regressed by the same frame-by-frame pseudo label, which is the clustering of \ac{MFCC} or hidden audio-visual representation in the transformer encoder. Thus, audio and visual context features are  synchronized.

Afterwards, a lip-reading network is constructed by combing the visual frontend and the transformer encoder pre-trained in self-supervision with a randomly initialized transformer decoder. This lip-reading network is fine-tuned under the supervision of text transcription. Once the fine-tuning is complete, the lip-reading network is frozen and serves as an expert in the training of talking face generation, as shown in \cref{fig:arch}.

\subsection{Audio encoder}
\label{subsec:input}
The audio encoder is to encode phoneme-level embeddings and provides the embeddings to the generator as the reference for mouth shape and lip movement.

\begin{figure}[!t]
    \centering
    \includegraphics[width=0.4\textwidth]{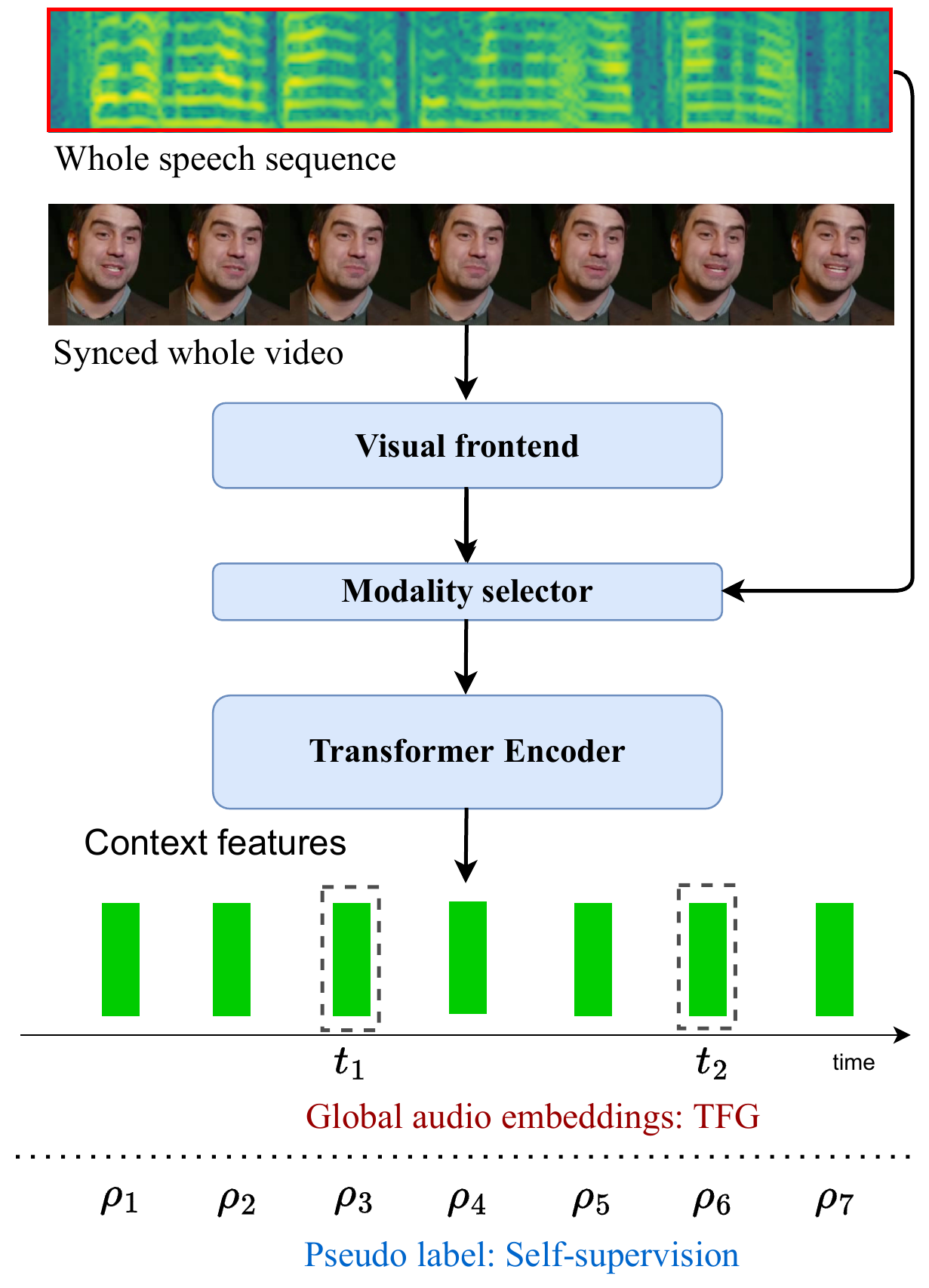}
    \caption{The self-supervised training of AV-Hubert and the process of capturing \textit{global} audio embeddings using the pre-trained transformer encoder. $\rho$ is frame-level pseudo labels for training.}
    \label{fig:transformer}
    \vspace{-5mm}
\end{figure}

We use two different audio encoders to extract the embeddings. We call them \textit{local} and \textit{global} audio embeddings.  The local audio embedding is extracted using a CNN-based network.
A 0.2-second audio segment whose centre is synchronized with the pose reference is fed as the input, as shown in \cref{fig:arch}. 
However, a 0.2-second audio segment does not contain temporal variations of the entire speech. The transformer encoder \cite{vaswani2017attention} is a popular method to address this limitation in some speech tasks \cite{ao-etal-2022-speecht5, tao2021someone}. Herein, we employ a pre-trained transformer encoder as mentioned in \cref{subsec:lipreading}, which is jointly trained with the lip-reading expert, as shown in \cref{fig:transformer}.
The transformer encoder takes the entire speech as input to generate audio context features of all frames. Then we choose a frame of the context features, which are time-aligned with the pose reference as the global audio embedding, as depicted in \cref{fig:transformer}.

\subsection{Video encoder}
The video encoder is to extract identity and pose information from images to a united visual embedding and supplies the embedding to the generator to synthesize an image consistent with provided identity and pose.

The visual embedding is extracted from two images: an identity reference and a pose reference. The pose reference is the same as the target face image, except it masks the lower-half face to prevent our TalkLip net from learning the movements in the lip region. The identity reference is randomly picked within the same video sequence. Two references are concatenated in the channel dimension as visual input and fed to a CNN-based video encoder.

\subsection{Video generation}
\label{subsec:vgen}

The generator constructs a talking face image given an audio embedding and a video embedding. As Unet \cite{ronneberger2015u} have a great trade-off between a reconstruction effect and computational efficiency \cite{li2016deep}, we adopt a skip connection like Unet between the video encoder and the generator, which is employed in \cite{prajwal2020lip} as well. The generator consists of transposed CNN blocks.

Given a synthesized talking face image and a ground truth image, two generation losses are applied to improve the video generation quality i.e., the reconstruction loss as \cref{eq:l1_loss} and the \ac{GAN} loss as \cref{eq:gen_loss}, \cref{eq:disc_loss}.

We adopt L1 as the reconstruction loss between the synthesized and real face images due to its better convergence to a local minimum than that of L2 loss \cite{zhao2016loss}. Given $N_i$ pairs of synthesized faces $v'$ and ground truth faces $v$, the reconstruction loss is formulated as:
\vspace{-2mm}
\begin{equation}
  \mathcal{L}_{\rm rec} = \frac{1}{N_i} \sum_i^{N_i} |v_i - v_i'|
  \label{eq:l1_loss}
\end{equation}
\vspace{-4mm}

 To increase the realism of synthesized talking face videos, many works adopt \ac{GAN}~\cite{park2022, liang2022expressive}. Considering its outstanding performance, in this paper, a \ac{GAN} loss is also applied. Let us denote $G$ and $D$ as the generator and discriminator in \cref{fig:arch}, respectively. To leverage \ac{GAN}s \cite{goodfellow2020generative}, two losses are involved as follows:
\vspace{-1mm}
\begin{equation}
  \mathcal{L}_{\rm gen} = \mathbb{E}[\log(1-D(v'))]
  \label{eq:gen_loss}
\end{equation}
\vspace{-4mm}
\begin{equation}
  \mathcal{L}_{\rm disc} = \mathbb{E}[\log(1-D(v))] + \mathbb{E}[\log(D(v'))]
  \label{eq:disc_loss}
\end{equation}
where $\mathcal{L}_{\rm disc}$ optimizes the discriminator $D$ to distinguish synthesized face image $v'$ from ground truth faces $v$, while $\mathcal{L}_{\rm gen}$ improves the quality of synthesized faces $v'$ to fool the discriminator.

\subsection{Lip reading loss}
\label{subsec:lip_loss}
The lip-reading expert is applied on hybrid videos i.e., the ground truth videos 
with some frames substituted by corresponding synthesized images. Therefore, the gradients on generated images will further back-propagate to optimize the generator, audio and video encoder.

Let us denote $V=[v_1, v_2', ..., v_T]\in\mathbb{R}^{T \times 3 \times H \times W}$ as a hybrid video, where $v$ and $v'$ are the real and synthesized images, respectively, and $T,H,W$ are the number of frames, height, and width. We follow the masking strategy in \cite{baevski2020wav2vec} to choose which frames of $v$ are replaced by $v'$. In particular, we randomly sample a proportion $p$ of all frames as a set of starting frames whose subsequent $M$ frames 
will be replaced with $v'$.
In this way, lip movements within $v'$ and between $v$ and $v'$ are considered. 
We assume $Y\in\mathbb{R}^{L \times C}$ as the text content of the ground truth video, where $L$ ad $ C$ denote the text length and the output class number. 

The lip-reading expert \cite{shi2022learning} contains (1) a 3D convolutional layer followed by a ResNet-18\cite{he2016deep} to extract features of lip movement; (2) a transformer encoder to produce a context feature $R\in\mathbb{R}^{T \times f}$ by calculating global temporal dependency where $f$ is the feature dimension and (3) a transformer decoder to predict the text  $\hat{Y}\in\mathbb{R}^{L \times C}$. Note that the lip-reading expert is frozen. Finally, we use cross-entropy  to compute the  lip-reading loss:
\begin{equation}
  \mathcal{L}_{\rm lip} = -Y\log P(\hat{Y}|V)
  \label{eq:lip_loss}
\end{equation}

\subsection{Contrastive loss}
\label{subsec:contrastive_loss}
We use the contrastive strategy to enhance lip-speech synchronization by attracting audio embeddings and their time-aligned visual context features while repelling the audio embeddings from different frames. To achieve such a goal, we select infoNCE~\cite{oord2018representation}.

Since the lip-reading expert is frozen, 
we do not use visual context features as anchors.
Instead, given audio embeddings $E^a$ from the audio encoder and visual context features $R$ from the lip-reading expert, the contrastive loss is computed as:
%
%
\begin{eqnarray}
  \label{eq:sim}
    \theta(x, x') \hspace{-0.2cm} &=& \hspace{-0.2cm}\exp(\mathcal{F}(x) \cdot \mathcal{F}(x')/\tau),\\
  \mathcal{L}_{\rm av} \hspace{-0.2cm}&=& \hspace{-0.2cm} -\sum_{i\in \Upsilon} \log\frac{\theta(E^a_i, R_i)}{\theta(E^a_i, R_i) + \mathop{\sum}_{
         j\in \Upsilon
        \atop
        i\neq j} \theta(E^a_i, E^a_j)}
\end{eqnarray}
where $\Upsilon$ denotes all masked frames described in \cref{subsec:lip_loss}. We follow \cite{khosla2020supervised} to employ Eq. (\ref{eq:sim}) to calculate the similarity of the two features, where $\tau$ is a pre-defined temperature parameter, $\mathcal{F}$ is a linear layer, and $x$ and $x'$ generally indicate two feature vectors. Besides, $x$ and $x'$ before and after $\mathcal{F}$ are normalized.


Finally, we combine 
all the aforementioned losses to optimize the network:
\begin{equation}
  \mathcal{L}_{t} = \lambda_l \cdot \mathcal{L}_{\rm lip} + \lambda_{c} \cdot \mathcal{L}_{\rm av}+ \lambda_{r} \cdot \mathcal{L}_{\rm rec} + \lambda_{g} \cdot \mathcal{L}_{\rm gan}
  \label{eq:total_loss}
\end{equation}
where $\lambda_l$, $\lambda_c$, $\lambda_r$ and $\lambda_g$ scale the contributions of different loss items. $\mathcal{L}_{\rm gan}$ is the sum of  $\mathcal{L}_{\rm gen}$ and $\mathcal{L}_{\rm disc}$.

\section{Experiments}\label{sec:exp}
\subsection{Experimental settings}
\textbf{Dataset} We train the proposed method in the train set (29 hours) of LRS2 and evaluate it with other competitive approaches in the test sets of two publicly available datasets: LRW \cite{chung2016lip}, and LRS2 \cite{afouras2018deep}.  LRW is a dataset for audio-visual word classification. Each sample in the LRW contains a video with 29 frames (about 1 second), synchronized audio, and a target word in the middle of the sample. There are 500 different words in the whole dataset. LRS2 is an open-world audio-visual speech recognition dataset. Each sample in it consists of video, audio and target text sequence. There are two subsets for training with a 60k vocabulary: pretrain (195 hours), and train (29 hours). Audio and video sample rates of both datasets are 16 Khz and 25 fps, respectively.

\textbf{Metrics} Comparisons are evaluated on the following widely used metrics: PSNR\cite{vougioukas2020realistic,park2022}, SSIM\cite{wang2004image, liang2022expressive}, LSE-C\cite{prajwal2020lip, zhou2021pose}. PSNR and SSIM are metrics to measure visual quality, which are calculated by comparing generated images with ground truth. LSE-C is proposed in \cite{prajwal2020lip} to quantify lip-speech synchronization through a pre-trained sync-net \cite{chung2017out}. The higher their scores, the better their performances.

Additionally, we propose a new strategy to evaluate reading intelligibility, including the Word Error Rate (WER) of synthesized video on LRS2, and the Word Accuracy (ACC) on the unseen dataset LRW. 
To evaluate the WER, we employ the AV-Hubert (large) \cite{shi2022learning}, which achieves \ac{SOTA} performance on the lip-reading task. We fine-tune it on the whole LRS2 dataset as one lip-reading observer. Its performance on the test set of LRS2 is 23.82\% \ac{WER}. Note that the lip-reading expert in our proposed method is different from the lip-reading observer since the lip-reading expert is AV-Hubert (base). 
Besides, To further increase the difference between lip-reading expert and observer, a lip-reading network adopting Conformer \cite{wang2022predict, ma2021end} serves as another lip-reading observer. We run Conformer in teacher-forcing mode \cite{williams1989learning} to calculate WER. In other words, ground truth text is involved in inference. For measuring ACC on LRW, we choose MVM \cite{kim2022distinguishing}, whose prediction accuracy on LRW is 88.5\%.

\textbf{Comparison methods} Five \ac{SOTA} methods of talking face generation are used to compare: ATVGnet \cite{chen2019hierarchical}, Wav2Lip \cite{prajwal2020lip}, Faceformer \cite{fan2022faceformer}, PC-AVS \cite{zhou2021pose}, SyncTalkFace \cite{park2022}. ATVGnet belongs to the intermediate representation-based method, which first predicts a face landmark and then generates a realistic image. Wav2Lip is a representative of reconstruction methods and claims \ac{SOTA} performance in lip sync synchronization. Faceformer represents an emerging manner, namely sequence-to-sequence manner, which employs a transformer to take the whole audio and output a video sequence. Originally, in \cite{fan2022faceformer}, this method is applied to generate a vertex sequence (3D scan data) from an audio sequence. It does not work if we simply replace the vertex sequence by the image sequence. Thus, we adopt the video encoder and corresponding visual input to replace the style encoder in \cite{fan2022faceformer} and add a skip connection between the visual input and the output image. PC-AVS is also a reconstruction method, which disentangles identity, speech content, and poses. SyncTalkFace addresses lip-speech synchronization via a memory network. The last two methods do not publicly share their code or training code. Thus, their performances are collected from \cite{park2022}.

\textbf{Implementation details} Below preprocessing is applied on raw audio and videos.  Following Wav2Lip \cite{prajwal2020lip}, we detect faces on each image as \ac{ROI} and then crop and resize the \ac{ROI} to 96$\times$96. Audio wavforms are preprocessed to mel-spectrogram with hop and window lengths, and mel bins are 12.5 ms, 50 ms, and 80.

We only fine-tune the last 3 transformer blocks of the pre-trained transformer audio encoder during the \ac{TFG} training. The detailed architectures of our method can be found in the supplementary. During inference, we always use the first image of a video as the identity reference. 

\begin{table*}[!htb]
\caption{Quantitative results on LRW and LRS2 (N.A. indicates the result is not applicable). Performances of methods with * are collected from \cite{park2022} which are trained using the whole LRS2 dataset (224 hours) while our methods are trained by the LRS2 dataset (29 hours). Thus, we can not evaluate their performances in reading intelligibility. WER$_1$, WER$_2$ are scored by the AV-Hubert (large) \cite{shi2022learning} and Conformer \cite{wang2022predict, ma2021end}. ACC is evaluated by MVM \cite{kim2022distinguishing}. 
$g$, $l$, and $c$ in parentheses behind \textbf{TalkLip} 
indicate global audio embedding, local audio embedding, and contrastive learning, respectively. \textbf{Base} refers to the \textbf{TalkLip} ($l$). }

\begin{tabular}{llllllllll}
\toprule
\multicolumn{1}{c}{\multirow{2}{*}{Method}} & \multicolumn{4}{c}{LRW}                & \multicolumn{5}{c}{LRS2}                            \\ \cline{2-10} 
\multicolumn{1}{c}{}                        & PSNR $\uparrow$ & SSIM $\uparrow$ & LSE-C $\uparrow$ & ACC (\%) $\uparrow$ & PSNR $\uparrow$ & SSIM $\uparrow$ & LSE-C $\uparrow$ & WER$_1$ (\%) & WER$_2$ $\downarrow$ \\
\midrule
Ground Truth                                &N.A.  &1.000   & 6.88  & 88.51   &N.A. &1.000   & 8.25  &  23.82     &  40.9         \\ \hdashline
ATVGnet                                     & 30.71  &0.791  & 5.64 & 18.10  & 30.42 & 0.751  &5.05 & 113.69        &  91.8   \\
Wav2Lip                                     &31.52  &0.874  & 7.18 & 59.98    &31.36 & 0.854 & 8.40 & 82.06 & 73.9 \\
Faceformer                                  &29.19  &0.856  & 5.58 & 53.43 &29.47& 0.840 & 6.42 & 97.64   & 79.0      \\
PC-AVS*                                     &30.44 &0.778   & 6.42  &  -       &29.89 & 0.747  & 6.73  &  -        &    -      \\
SyncTalkFace*                               &\textbf{33.13} &\textbf{0.893}   & 6.62  &  -       &\textbf{32.59} & \textbf{0.876}  & 7.93  &  -        &    -      \\
\hdashline
\textbf{TalkLip} ($l$)   &31.24  &0.867   & 6.44  & 79.78 & 31.38 & 0.849    & 7.58  &  45.74    &  55.7   \\ 
\textbf{TalkLip} ($l$ + $c$) &31.52  &0.867   & 6.51  & 83.17 & 31.14 & 0.850   & 7.76  &  38.00     & 49.2          \\
\textbf{TalkLip} ($g$)    &30.78  &0.871   & 7.01  & 86.57  & 30.86 & 0.854   & 8.38  &  25.31   &  36.5   \\ 
\textbf{TalkLip} ($g$ + $c$)  &31.18  &0.866  & \textbf{7.28}  & \textbf{87.81}  & 31.19 &  0.850   & \textbf{8.53}  & \textbf{23.43}  &  \textbf{35.1}    \\
\hdashline
\textbf{Base} w.o.$\mathcal{L}_{\rm lip}$   & 31.22  & 0.865  & 6.01  & 48.58  & 31.08 &  0.852   & 7.09  & 103.57  & 82.2  \\
\textbf{Base} w.o.$\mathcal{L}_{\rm lip, gan}$     & 30.64 & 0.864  & 5.03  & 30.80  & 30.70 &  0.851   & 5.93  & 116.26  & 89.3    \\

\bottomrule
\end{tabular}
\label{tab:quan}
\end{table*}

\subsection{Experimental result}
\textbf{Quantitative results}
Performance comparisons on visual quality, lip-speech sync and reading intelligibility are shown in \cref{tab:quan}. LRS2 and unseen LRW are involved. With the help of the global audio encoder and contrastive learning, the TalkLip ($g$ + $c$) achieves the \ac{SOTA} performance on lip-speech sync. Although Wav2Lip is superior to other TalkLip nets in lip-speech sync (LSE-C), its performance in reading intelligibility (WER$_1$ and WER$_2$) on LRS2 lags behind all the TalkLip nets. This suggests that improving lip-speech sync does not lead to better reading intelligibility. 

On LRS2, it is observed that our methods outperform others in reading intelligibility while they are close to or even better than real videos. Such advantage remains in the unseen LRW dataset, which suggests that our method is robust across different datasets and lip-reading tasks e.g., lip-reading of a continuous utterance and a single word. We note that the TalkLip ($g$ + $c$) even has better reading intelligibility than the ground truth. We hypothesize that the lip-reading expert is trained to capture relative standard and full lip movements. And the audio transformer encoder is trained well to extract phoneme-level information. Thus TalkLip nets learn from the lip-reading expert and tend to generate more standard lip movement than natural lip movements. Note that WER$_1$ of some methods are higher than 100\%. The reason is that the lip-reading observer of WER$_1$ does not know the length of the target text and may predict a much longer sentence. In this case, the edit distance algorithm can output a WER excessing 100\%. Differently, the teacher-forcing mode knows the length so that WER$_2$ $<$ 100\%.

PSNR and SSIM show that the visual quality of our methods is worse than SyncTalkFace. We note that SyncTalkFace is trained on the whole LRS2 dataset, whose training data is 8 times more than ours. Despite the limited training data, our results are close to the second-best method, Wav2Lip. 

\begin{figure*}
      \begin{subfigure}{0.54\textwidth}\label{subfig:displaya}
    \includegraphics[width=9.38cm]{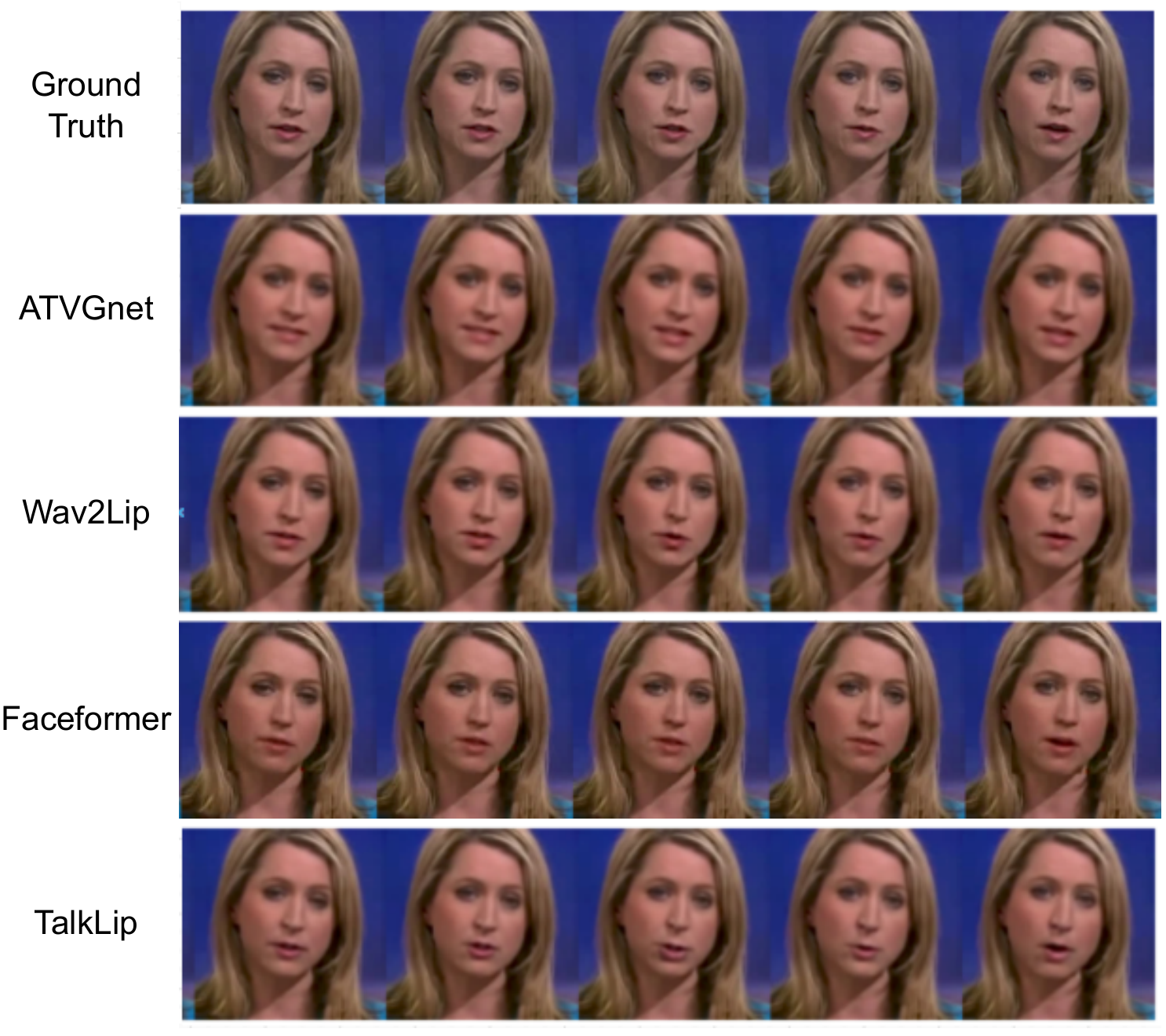}
    \caption{\textit{`Another two \underline{hundreds}'}}
   \end{subfigure}
         \begin{subfigure}{0.45\textwidth}\label{subfig:displayb}
    \includegraphics[width=8cm]{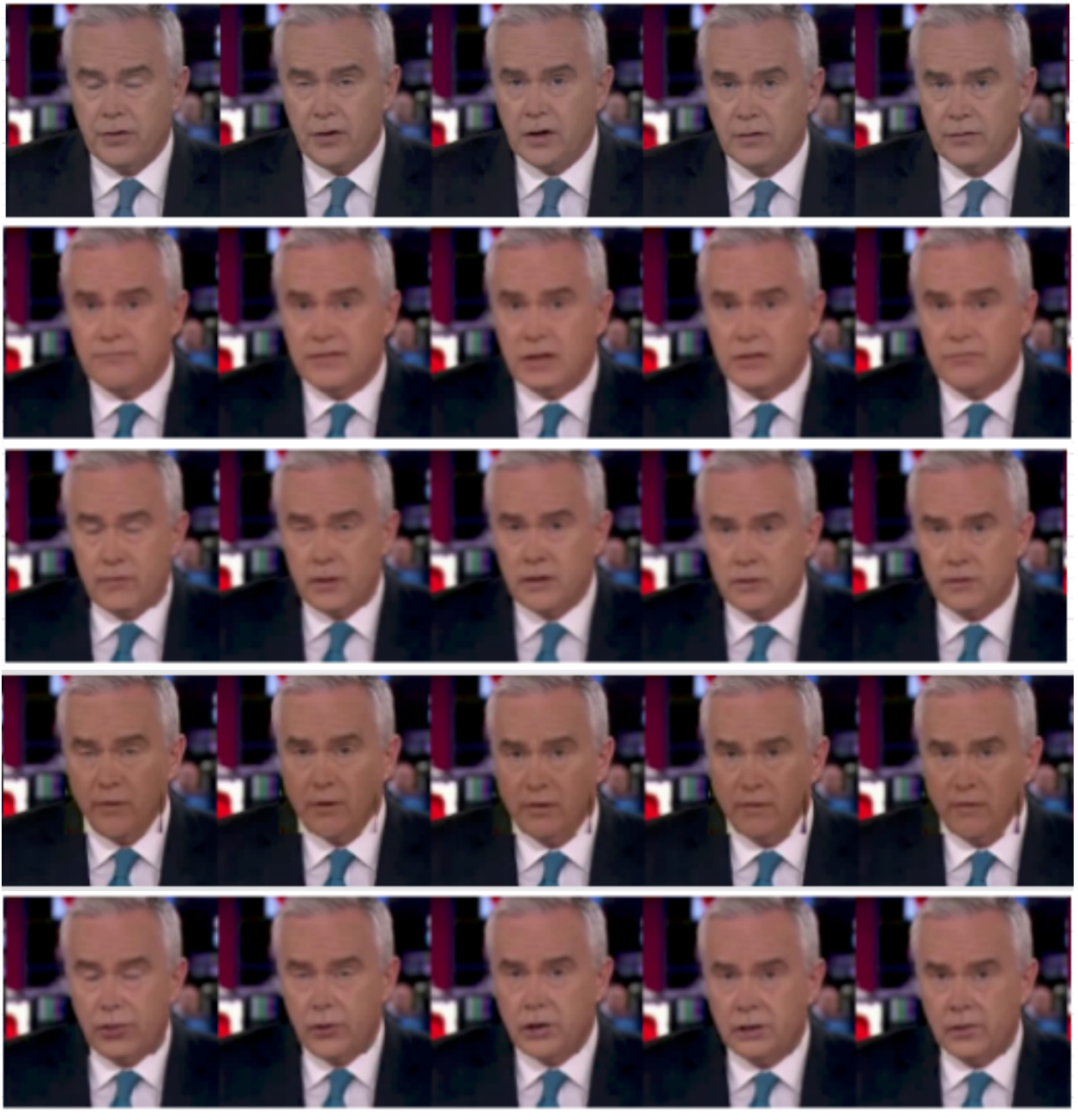}
    \caption{\textit{`More than a \underline{third} of the world'}}
   \end{subfigure}
    \vspace{-2mm}
    \caption{Snapshots of the generated talking face videos correspond to a part of the underlined text.
    }
    \label{fig:display}
    \vspace{-2mm}
\end{figure*}

\textbf{Qualitative results}
To qualitatively evaluate different methods, we provide the uniformly sampled snapshots of two generated talking face videos in Fig.~\ref{fig:display}. Specifically, the ground truth video is provided in the first row where synthesized images of different methods follow the next.
From Fig.~\ref{fig:display} (a), we can see that our proposal provides the most similar image frames as the ground truth video with smooth variations in mouth shape. For Fig.~\ref{fig:display} (b), we are also of the best performance in video quality and the naturalness in mouth movements.


\begin{table}[]
\caption{User study on face videos  generated by different methods in visual quality (VQ), lip-speech sync (Sync), realness (Real), and reading intelligibility (RI).}
\centering
\begin{tabular}{lcccc}
\toprule
Method        & VQ $\uparrow$ & Sync $\uparrow$ & Real $\uparrow$   & RI $\uparrow$   \\ \hline
ATVGnet \cite{chen2019hierarchical}  &  3.27       & 2.96        &    2.73        &  3.20     \\
Wav2Lip \cite{prajwal2020lip} &   3.75      &    3.77     &      3.75      &    3.38   \\
Faceformer \cite{fan2022faceformer}        &   2.84      &    2.77     &   2.56         &    3.04   \\
TalkLip ($g$ + $c$) &   \textbf{3.78}      &   \textbf{4.00}      &     \textbf{3.89}       & \textbf{3.76} \\    
\bottomrule
\end{tabular}
\label{tab:userstudy}
\vspace{-3mm}
\end{table}


\begin{figure}
  \centering
      \begin{subfigure}{0.49\columnwidth}
    \includegraphics[width=4.5cm]{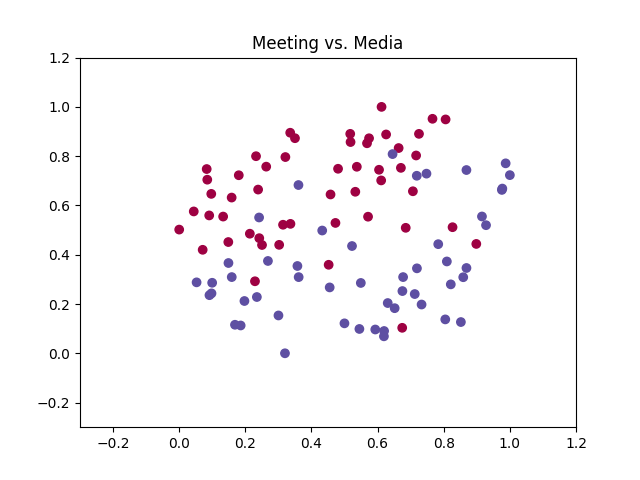}
        \caption{TalkLip ($l$)}
  \end{subfigure}
    \begin{subfigure}{0.49\columnwidth}
    \includegraphics[width=4.5cm]{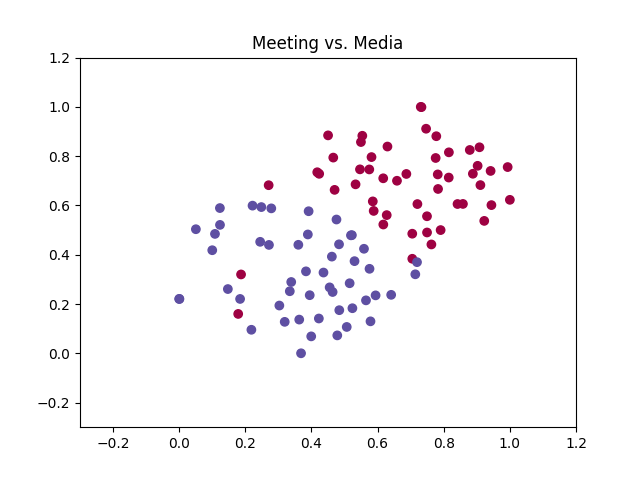}
    \caption{TalkLip ($l$+$c$)}
  \end{subfigure}
    \caption{The t-SNE visualization of audio embeddings correspond to `Meeting' (red) and `Media' (blue).}
    \vspace{-4mm}
    \label{fig:distri_audio_mm}
\end{figure}




\subsection{User study}
We invite 15 people to evaluate the generated talking face videos of our method 
with the other 3 \ac{SOTA} methods. 6 videos from LRS2 and 9 videos from LRW are involved in this user study with human measures in visual quality, speech-lip sync and realness. Each metric ranges from 1 to 5. 
Moreover, to evaluate reading intelligibility, we randomly collect 5 videos from LRW for participants to score the reading intelligibility of the corresponding ground-truth words. We do not use samples from LRS2 since lip reading on continuous long videos is hard for people without professional lip-reading training. The results from all participants are averaged and listed in \cref{tab:userstudy}.

From the results, we can see that our proposal achieves the best performance in all subjective measures. 
Specifically, for visual quality, the results are consistent with that in \cref{tab:quan}. The performance of our methods approximates Wav2Lip and is better than others.
For lip-speech synchronization and realism, our methods outperform others by a distinguishable margin.
This verifies the capability of our method in generating videos of human naturalness. Finally, for reading intelligibility of the generated face videos, our proposal obtains the score of 3.76, beats the second-best method i.e., Wav2Lip by 0.38 (11.2\%) and the worst method i.e., Faceformer by 0.72 (23.7\%) thanks to the assistance of the lip-reading expert.

\subsection{Ablation study on the GAN loss}
Although the GAN loss is designed to make synthesized videos more realistic, we find it also lead to better lip-speech sync and reading intelligibility. Compared with \textbf{Base} {w.o. $\mathcal{L}_{\rm lip, gan}$}, \textbf{Base} {w.o. $\mathcal{L}_{\rm lip}$} improves LSE-C by about 19.5 \% on LRW and LRS2. Besides, reading intelligibility is also improved, especially on LRW.

\subsection{Ablation study on the lip-reading loss}

We propose to employ a lip-reading expert to improve reading intelligibility of synthesized talking faces. Compared with the TalkLip ($l$) without the lip-reading loss, the TalkLip ($l$) with the lip-reading loss shows at least 26.5\% improvement on LRS2 and 31.2\% improvement on LRW. These advantages confirm the positive effect of the lip-reading loss on improving reading intelligibility.

\subsection{Ablation study on contrastive learning}
We propose contrastive learning to enhance lip-speech synchronization. From \cref{tab:quan}, the contribution of contrastive learning is proven. LSE-C of the TalkLip ($l$ + $c$) is higher than that of the TalkLip ($l$) on both LRS2 and LRW datasets. The only difference between them is the former employs contrastive learning. Besides, the TalkLip ($g$ + $c$) also has this advantage over the TalkLip ($g$).

Moreover, it is observed that reading intelligibility is also improved with contrastive learning. In particular, as shown in \cref{tab:quan}, the TalkLip ($l$ + $c$) gains an improvement of at least 6.5\% \ac{WER} on LRS2 and 3.39\% ACC on LRW by involving contrastive learning. The improvement may benefit from the fact that contrastive learning pushes audio embeddings from different frames more apart. To validate this guess, we pick a pair of words in LRW and visualize the distributions of audio embeddings using t-SNE~\cite{van2008visualizing} in \cref{fig:distri_audio_mm}. In detail, LRW provides the start and end times of  ground-truth words. Thus, for each sample, we calculate audio embeddings at the ground-truth interval and sum them up as the feature to be visualized. The pair are `Meeting' and `Media' which have a similar prefix. 


It is shown that audio embeddings of `Meeting' and `Media' are mixed without contrastive learning, while there are only three outliers with contrastive learning.

With these results, we conclude that contrastive learning is capable of enhancing lip-speech synchronization and repelling audio embeddings at different frames.

\subsection{Ablation study on global audio embedding}
We propose to use global audio embedding because of two merits: (1) global audio embeddings are calculated from entire utterances by a transformer encoder, which benefits from the global temporal dependency to represent phoneme-level information and (2) the transformer encoder is pre-trained synchronically with the lip-reading expert and therefore its outputs are synchronized with visual content features. 

From \cref{tab:quan}, the first merit is proven because the TalkLip ($g$ + $c$) outperforms the TalkLip ($l$ + $c$) both in \ac{WER} on LRS2 and ACC on LRW. 
It is observed that two TalkLip nets with the global audio encoder outperform those with the local audio encoder on LSE-C, as shown in \cref{tab:quan}. This improvement confirms the benefit of audio-visual synchronical pre-training to lip-speech synchronization.


\section{Conclusion}\label{sec:conclusion}
In this paper, we analyse the importance of reading intelligibility and propose a novel TalkLip net to synthesize talking face videos with great reading intelligibility. Particularly, a TalkLip net employs a pre-trained lip-reading expert to penalize the incorrect lip-reading predictions from synthesized videos. To evaluate reading intelligibility, we propose a new strategy that contains single-word ACC measurement and \ac{WER} on videos of continuous utterances and prepares a benchmark code. From the results, TalkLip nets are superior to the other competitive methods on reading intelligibility by a large margin both subjectively and objectively. Besides, based on the lip-reading expert, we propose a novel contrastive learning to enhance lip-speech synchronization, and a synchronically pre-trained audio transformer encoder to leverage global temporal dependency. Our extensive experiments confirm that the contrastive learning is effective to lip-speech synchronization while it also improves the reading intelligibility. Our results also prove that the audio transformer encoder generates better phoneme-level audio embeddings and helps lip-speech synchronization. Therefore, our TalkLip net can synthesize talking face videos with superior reading intelligibility, lip-speech synchronizaiton and competitive visual quality. 

\noindent\textbf{Acknowledgement} This work is supported by 1) Huawei Noah's Ark Lab; 2) National Natural Science Foundation of China (Grant No. 62271432); 3) Internal Project Fund from Shenzhen Research Institute of Big Data (Grant No. T00120220002); 4) Guangdong Provincial Key Laboratory of Big Data Computing, The Chinese University of Hong Kong, Shenzhen (Grant No. B10120210117-KP02); 5) Human-Robot Collaborative AI for Advanced Manufacturing
and Engineering (Grant No. A18A2b0046),  Agency of Science, Technology and Research (A*STAR), Singapore. 6) MOE AcRF Tier, A-0009455-01-00.


{\small
\bibliographystyle{IEEEtran}
\bibliography{egbib}
}

\newpage

\section{Supplementary}
\subsection{Audio Encoder Architecture}
\label{sec:aud_enc}

The details of the audio encoder to extract a local audio embedding given a spectrogram of a 0.2-second audio segment are shown in \cref{tab:aud_enc}.
Specifically, layers between two lines compose a block whose input and output  are summed up via a residual connection. Meanwhile, batch normalization is applied after each of the convolutional layers.

\begin{table}[htb]
\centering
  \caption{Audio encoder architecture. All parameters listed in the `Filters' column are kernel sizes, output channels, strides, padding, and repetition of layers.}
\begin{tabular}{lll}
\toprule
Layer Type & Filters                    & Output dim.  \\ 
\midrule
Conv 2D    & \{[3, 3], 32, [1, 1], 1\} $\times$ 1    & 32$\times$80$\times$16  \\
Conv 2D    & \{[3, 3], 32, [1, 1], 1\} $\times$ 2    & 32$\times$80$\times$16          \\
\midrule
Conv 2D    & \{[3, 3], 64, [3, 1], 1\} $\times$ 1  &  64$\times$27$\times$16      \\
Conv 2D    & \{[3, 3], 64, [1, 1], 1\} $\times$ 2  &   64$\times$27$\times$16            \\
\midrule
Conv 2D    & \{[3, 3], 128, [3, 3], 1\} $\times$ 1  &  128$\times$9$\times$6             \\
Conv 2D    & \{[3, 3], 128, [1, 1], 1\} $\times$ 2 &   128$\times$9$\times$6   \\
\midrule
Conv 2D    & \{[3, 3], 256, [3, 2], 1\} $\times$ 1  & 256$\times$3$\times$3            \\
Conv 2D    & \{[3, 3], 256, [1, 1], 1\} $\times$ 1 &   256$\times$3$\times$3                  \\
\midrule
Conv 2D    & \{[3, 3], 512, [1, 1], 0\} $\times$ 1 &   512$\times$1$\times$1             \\
Conv 2D    & \{[1, 1], 512, [1, 1], 0\} $\times$ 1 & 512$\times$1$\times$1       \\
\bottomrule
\end{tabular}
  \label{tab:aud_enc}
\end{table}

\subsection{Audio Transformer Encoder Architecture}
\label{sec:aud_t_enc}

The audio transformer encoder is used to extract phoneme-level information in speech considering the global temporal dependency, namely global audio embeddings. We use speeches with varying lengths as the input to the transformer encoder. In practice, a speech is preprocessed to a spectra $S\in\mathbb{R}^{T \times F}$, where $T$ and $F$ are the numbers of frames and filter banks. Then, every 4 frames are stacked into one frame. Herein, $F$ is fixed to 26.

There are 12 cascaded transformer blocks in the transformer encoder. The hidden layer dimension, feed-forward layer dimension and the number of attention heads are set to 768, 3072 and 12, respectively. Thus, the output of the transform encoder is denoted as $Z\in\mathbb{R}^{(T/4) \times 768}$. Afterwards, we take one frame of $Z$, which is timely aligned with the pose reference, as the global audio embedding.

\subsection{Video Encoder Architecture}
\label{sec:vid_enc}

We use a video encoder to extract the identity and pose information to a united visual embedding from a concatenation (6$\times$96$\times$96) of an identity and a pose image. Tab. \ref{tab:vid_enc} illustrates the detailed architecture.

\begin{table}[htb]
\centering
  \caption{Video encoder architecture. All parameters listed in the `Filters' column are kernel sizes, output channels, strides, padding, and repetition of layers.}
\begin{tabular}{lll}
\toprule
Layer Type & Filters                    & Output dim.  \\ 
\midrule
Conv 2D    & \{[7, 7], 16, [1, 1], 3\} $\times$ 1    & 16$\times$96$\times$96  \\
\midrule
Conv 2D    & \{[3, 3], 32, [2, 2], 1\} $\times$ 1    & 32$\times$48$\times$48          \\
Conv 2D    & \{[3, 3], 32, [1, 1], 1\} $\times$ 2    & 32$\times$48$\times$48          \\
\midrule
Conv 2D    & \{[3, 3], 64, [2, 2], 1\} $\times$ 1  &  64$\times$24$\times$24      \\
Conv 2D    & \{[3, 3], 64, [1, 1], 1\} $\times$ 3  &   64$\times$24$\times$24            \\
\midrule
Conv 2D    & \{[3, 3], 128, [2, 2], 1\} $\times$ 1  &  128$\times$12$\times$12             \\
Conv 2D    & \{[3, 3], 128, [1, 1], 1\} $\times$ 2 &   128$\times$12$\times$12   \\
\midrule
Conv 2D    & \{[3, 3], 256, [2, 2], 1\} $\times$ 1  & 256$\times$6$\times$6            \\
Conv 2D    & \{[3, 3], 256, [1, 1], 1\} $\times$ 2 &   256$\times$6$\times$6                  \\
\midrule
Conv 2D    & \{[3, 3], 512, [2, 2], 1\} $\times$ 1 &   512$\times$3$\times$3             \\
Conv 2D    & \{[3, 3], 512, [1, 1], 1\} $\times$ 1 & 512$\times$3$\times$3      \\
\midrule
Conv 2D    & \{[3, 3], 512, [1, 1], 0\} $\times$ 1 & 512$\times$1$\times$1             \\
Conv 2D    & \{[1, 1], 512, [1, 1], 0\} $\times$ 1 & 512$\times$1$\times$1      \\
\bottomrule
\end{tabular}
  \label{tab:vid_enc}
\end{table}

\subsection{Generator Architecture}
\label{sec:gen}

The details of the generator to synthesize a face image based on the concatenated audio and video embedding are provided  in Tab. \ref{tab:gen}:

\begin{table}[htb]
\vspace{-2mm}
\centering
  \caption{Generator architecture. All parameters listed in the `Filters' column for Conv2D are kernel sizes, output channels, strides, padding, and repetition of layers. Conv 2D T. means 2D transposed convolutional layers which has an extra parameter called output padding, placed after the padding parameter.  }
\begin{tabular}{lll}
\toprule
Layer Type & Filters                    & Output dim.  \\ 
\midrule
Conv 2D     & \{[1, 1], 512, [1, 1], 0\} $\times$ 1    & 512$\times$1$\times$1  \\
\midrule
Conv 2D T.   & \{[3, 3], 512, [2, 2], 0, 0\} $\times$ 1    & 512$\times$3$\times$3          \\
Conv 2D    & \{[3, 3], 512, [1, 1], 1\} $\times$ 1    & 512$\times$3$\times$3          \\
\midrule
Conv 2D T.    & \{[3, 3], 512, [2, 2], 1, 1\} $\times$ 1  &  512$\times$6$\times$6      \\
Conv 2D    & \{[3, 3], 512, [1, 1], 1\} $\times$ 2 &   512$\times$6$\times$6            \\
\midrule
Conv 2D T.   & \{[3, 3], 384, [2, 2], 1, 1\} $\times$ 1  &  384$\times$12$\times$12     \\
Conv 2D    & \{[3, 3], 384, [1, 1], 1\} $\times$ 2 &   384$\times$12$\times$12   \\
\midrule
Conv 2D T.    & \{[3, 3], 256, [2, 2], 1, 1\} $\times$ 1  & 256$\times$24$\times$24            \\
Conv 2D    & \{[3, 3], 256, [1, 1], 1\} $\times$ 2 &   256$\times$24$\times$24                 \\
\midrule
Conv 2D T.   & \{[3, 3], 128, [2, 2], 1, 1\} $\times$ 1 &   128$\times$48$\times$48             \\
Conv 2D    & \{[3, 3], 128, [1, 1], 1\} $\times$ 2 & 128$\times$48$\times$48      \\
\midrule
Conv 2D T.   & \{[3, 3], 64, [2, 2], 1, 1\} $\times$ 1 & 64$\times$96$\times$96             \\
Conv 2D    & \{[1, 1], 64, [1, 1], 1\} $\times$ 2 & 64$\times$96$\times$96      \\
\midrule
Conv 2D    & \{[3, 3], 32, [1, 1], 1\} $\times$ 1 & 32$\times$96$\times$96             \\
Conv 2D    & \{[1, 1], 3, [1, 1], 0\} $\times$ 1 & 3$\times$96$\times$96      \\
\bottomrule
\end{tabular}
  \label{tab:gen}
\vspace{-3mm}
\end{table}

Besides, the skip connection like Unet \cite{ronneberger2015u, prajwal2020lip} is applied. Particularly, hidden features in the generator are concatenated with hidden features in the video encoder with the same shape.

\subsection{Discriminator Architecture}
\label{sec:dis}

The details of the discriminator to penalize unrealistic synthesized face images are provided in Tab. \ref{tab:dis}. The discriminator only takes the lower half of faces as inputs.

\begin{table}[htb]
\centering
  \caption{Audio encoder architecture. All parameters listed in the `Filters' column are kernel sizes, output channels, strides, padding, and repetition of layers.}
\begin{tabular}{lll}
\toprule
Layer Type & Filters                    & Output dim.  \\ 
\midrule
Conv 2D    & \{[7, 7], 32, [1, 1], 3\} $\times$ 1    & 32$\times$48$\times$96  \\
\midrule
Conv 2D    & \{[5, 5], 64, [1, 2], 1\} $\times$ 1  &  64$\times$48$\times$48      \\
Conv 2D    & \{[5, 5], 64, [1, 1], 2\} $\times$ 1  &   64$\times$48$\times$48            \\
\midrule
Conv 2D    & \{[5, 5], 128, [2, 2], 2\} $\times$ 1  &  128$\times$24$\times$24             \\
Conv 2D    & \{[5, 5], 128, [1, 1], 2\} $\times$ 1 &   128$\times$24$\times$24   \\
\midrule
Conv 2D    & \{[5, 5], 256, [2, 2], 2\} $\times$ 1  & 256$\times$12$\times$12          \\
Conv 2D    & \{[5, 5], 256, [1, 1], 2\} $\times$ 1 &   256$\times$12$\times$12                  \\
\midrule
Conv 2D    & \{[5, 5], 512, [2, 2], 2\} $\times$ 1  & 512$\times$6$\times$6          \\
Conv 2D    & \{[5, 5], 512, [1, 1], 2\} $\times$ 1 &   512$\times$6$\times$6                    \\
\midrule
Conv 2D    & \{[3, 3], 512, [2, 2], 1\} $\times$ 1 &   512$\times$3$\times$3             \\
Conv 2D    & \{[3, 3], 512, [1, 1], 1\} $\times$ 1 & 512$\times$3$\times$3       \\
\midrule
Conv 2D    & \{[3, 3], 512, [1, 1], 0\} $\times$ 1 &   512$\times$1$\times$1             \\
Conv 2D    & \{[1, 1], 512, [1, 1], 0\} $\times$ 1 & 512$\times$1$\times$1       \\
\bottomrule
\end{tabular}
  \label{tab:dis}
\end{table}

\subsection{Qualitative Ablation Study on Contrastive Learning}
\label{sec:Contra}

Our experiments have confirmed that contrastive learning is effective in lip-speech synchronization, which also improves reading intelligibility. In this section, we visualize audio embeddings of the pairs of `Around' and `Ground' which is one of the most frequently confused word pairs \cite{chung2016lip}.

\begin{figure}
  \centering
      \begin{subfigure}{0.49\columnwidth}
    \includegraphics[width=4.5cm]{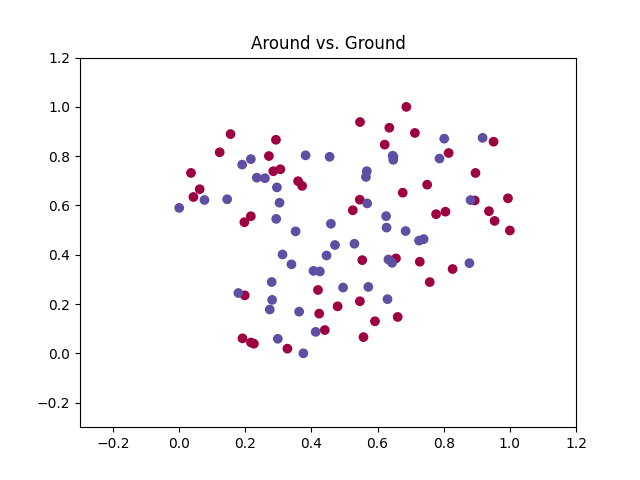}
    \caption{TalkLip ($l$)}
  \end{subfigure}
    \begin{subfigure}{0.49\columnwidth}
    \includegraphics[width=4.5cm]{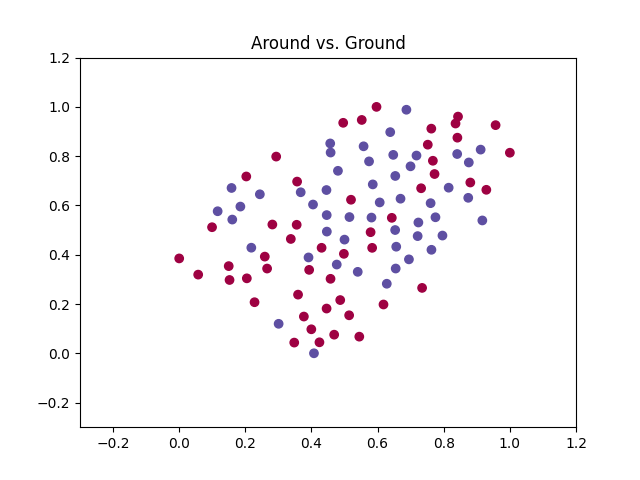}
    \caption{TalkLip ($l$+$c$)}
  \end{subfigure}
  \caption{The t-SNE visualization of audio embeddings correspond to  `Around' (red) and `Ground' (blue).}
    \label{fig:distri_audio_ag}

\end{figure}

As shown in \cref{fig:distri_audio_ag}, although audio embeddings of two words by the TalkLip ($l$ + $c$) are still not separated, the lower half of Fig. \textcolor{red}{1b} is mainly composed by red points, which is much better than the TalkLip ($l$).

\begin{figure}[!t]
    \centering
    \includegraphics[width=0.45\textwidth]{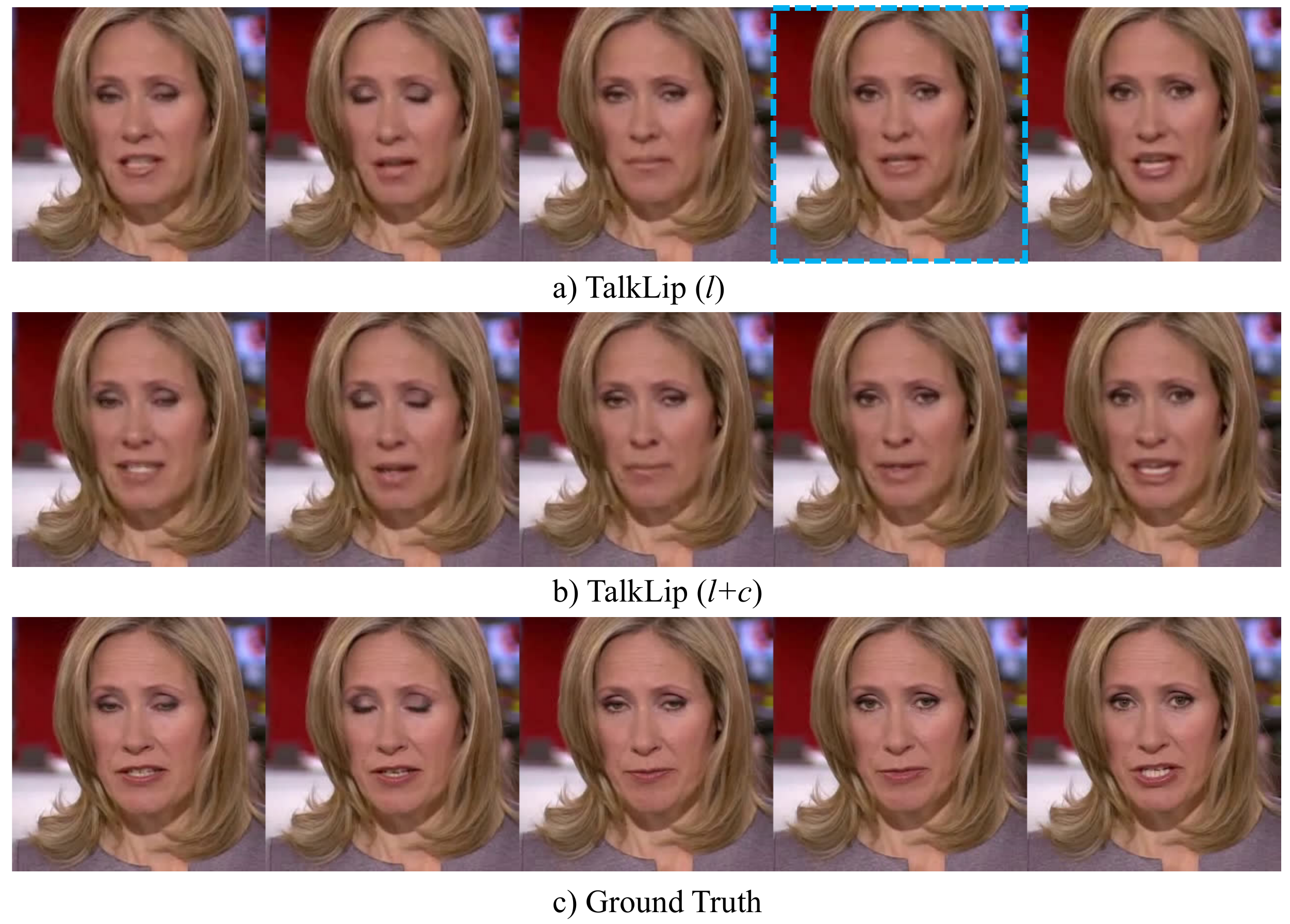}
    \vspace{-2mm}
    \caption{Snapshots of the generated talking face videos to demonstrate the benefit of the contrastive learning.}
    \label{fig:ablation_cont}
    \vspace{-3mm}
\end{figure}

Besides, we provide some figures to conduct a qualitative analysis. As shown in \cref{fig:ablation_cont}, it is observed that the TalkLip ($l+c$) is better than Prop. ($l$) with the help of contrastive learning. Especially the fourth image (in the blue box) of the TalkLip ($l$) is a little ahead of the ground truth. The fourth image of the TalkLip ($l+c$) is more synchronized than that of the TalkLip ($l$) with the ground truth.  

\subsection{Qualitative Ablation Study on Global Audio Embedding}
\label{sec:global}

\begin{figure}[!t]
    \centering
    \includegraphics[width=0.35\textwidth]{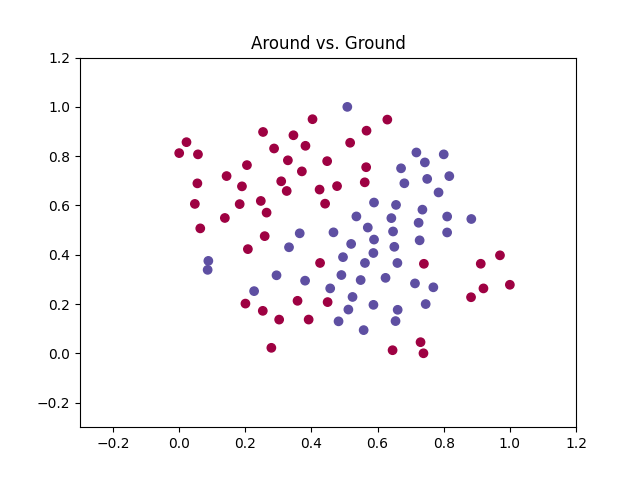}
    \caption{The t-SNE visualization of global audio embeddings correspond to `Around' (red) and `Ground' (blue).}
    \label{fig:v2tse}
\end{figure}

\begin{figure}[!t]
    \centering
    \includegraphics[width=0.45\textwidth]{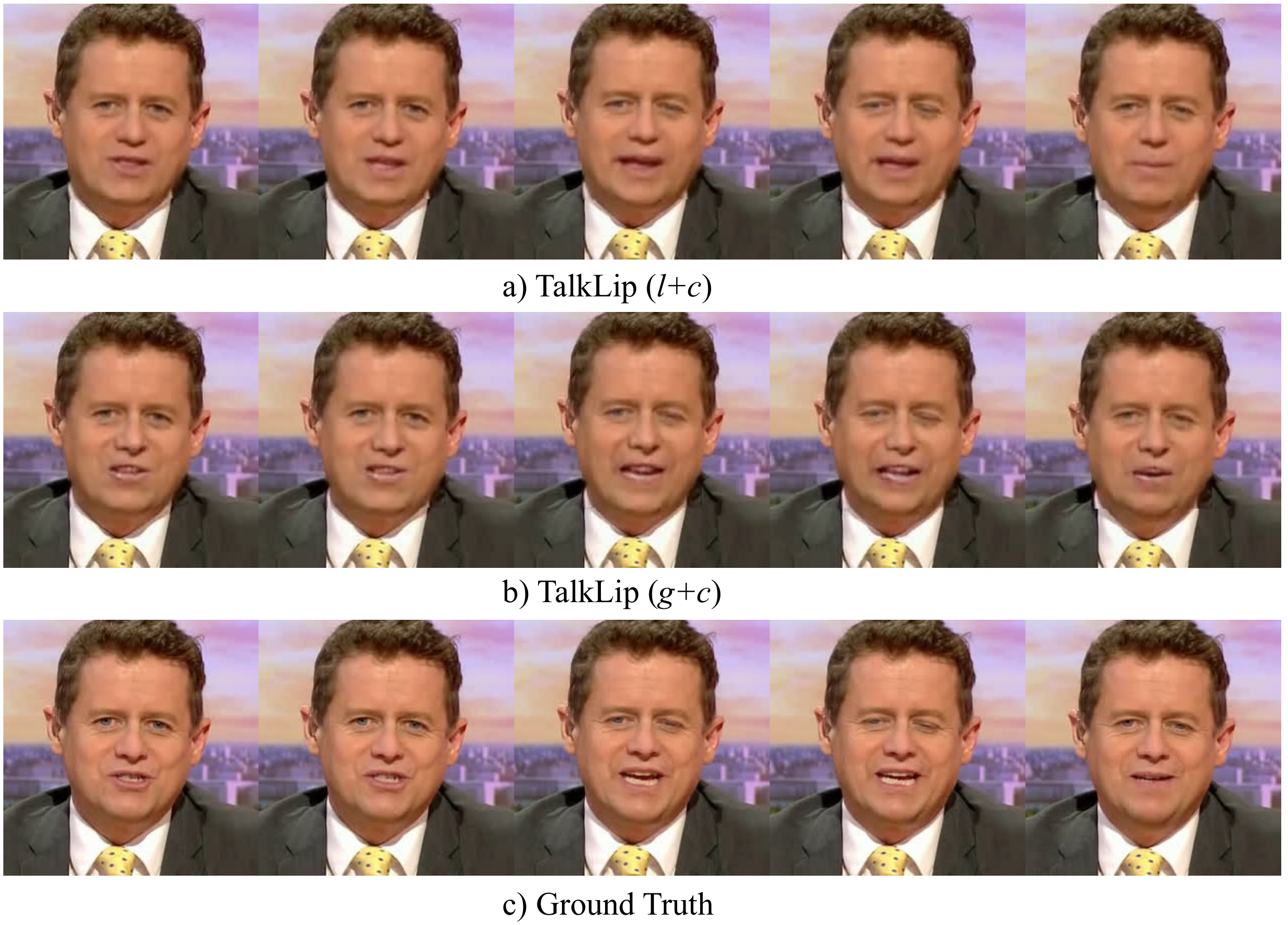}
      \vspace{-2mm}
    \caption{Snapshots of the generated talking face videos to demonstrate the benefit of using the transformer encoder which extracts the global audio embedding.}
    \label{fig:ablation_trans}
    \vspace{-3mm}
\end{figure}

To show a better representation of global audio embeddings in phoneme-level information, we visualize their distributions of `Around' and `Ground' in \cref{fig:v2tse}. It is observed that global audio embeddings are more separable than local audio embeddings as shown in \cref{fig:distri_audio_ag}. Besides, We show an image comparison between the TalkLip ($g+c$) and the TalkLip ($l+c$) as \cref{fig:ablation_trans}. It is observed that the lip movement of the TalkLip ($g+c$) is fuller than the TalkLip ($l+c$), which confirms the benefit of the global audio embedding.

\subsection{LSE-D}

LSE-D \cite{prajwal2020lip} is another metric to measure lip-speech synchronization. We provide a comparison of LSE-D in \cref{tab:quan}. We can observe that LSE-D also confirms the \ac{SOTA} performance of TalkLip ($g$ + $c$) on lip-speech synchronization.

\begin{table}[] %
\centering
\caption{LSE-D results on LRW and LRS2. Performances of methods with * are collected from \cite{park2022} which are trained using the whole LRS2 dataset (224 hours) while our methods are trained by the LRS2 dataset (29 hours).  }
\begin{tabular}{ccc}
\toprule
           & LRW                & LRS2    \\                  
\midrule
Ground Truth          & 6.97  & 6.45  \\ 
\hdashline
ATVGnet               & 8.56  & 8.65 \\
Wav2Lip               & 7.01  & 6.58  \\
Faceformer            & 8.00  & 7.80   \\
PC-AVS*               & 7.34 & 7.30   \\
SyncTalkFace*         &  6.97  &  6.26  \\
\hdashline
\textbf{TalkLip} ($l$)   & 7.00  & 6.63   \\ 
\textbf{TalkLip} ($l$ + $c$) & 7.00  & 6.56    \\
\textbf{TalkLip} ($g$)    & 6.75  & 6.01  \\ 
\textbf{TalkLip} ($g$ + $c$)  & \textbf{6.51} & \textbf{6.00}   \\
\bottomrule
\end{tabular}
\label{tab:quan}
\end{table}

\subsection{Limitation}

In the Fig. \textcolor{red}{1} of the main body, it is observed that our methods do not help improve visual quality. The lip-reading loss does not direct a better visual quality since PSNR and SSIM of the TalkLip ($l$) and the \textbf{Base} w.o. $\mathcal{L}_{\rm lip}$ are very close. Contrastive learning and the global audio encoder also do not boost visual quality as all four TalkLip nets show similar PSNR and SSIM. We will explore methods of improving visual quality in our further work.

\subsection{Qualitative Result}
\label{sec:qr}

\begin{figure}[!htb]
  \centering
  \begin{subfigure}{\columnwidth}
    \centering
    \includegraphics[width=7.5cm]{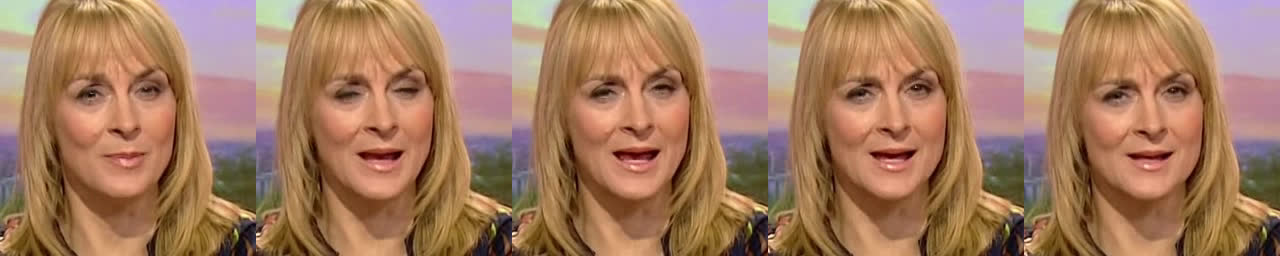}
    \caption{Ground Truth}
  \end{subfigure}
  \begin{subfigure}{\columnwidth}
      \centering
    \includegraphics[width=7.5cm]{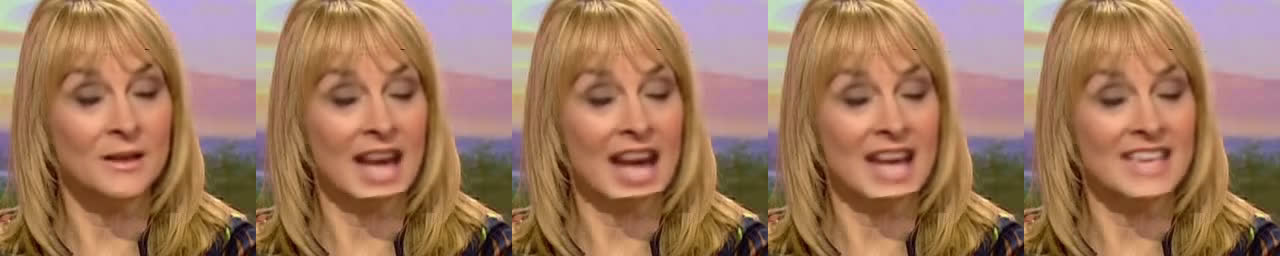}
    \caption{ATVGnet}
  \end{subfigure}
  \begin{subfigure}{\columnwidth}
      \centering
    \includegraphics[width=7.5cm]{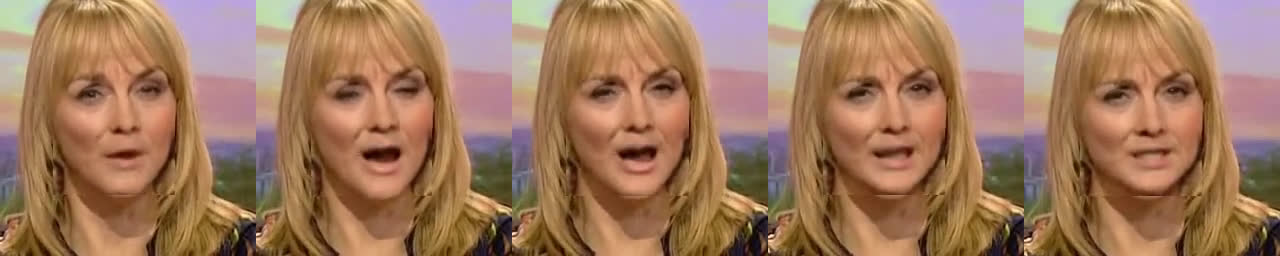}
    \caption{Wav2Lip)}
  \end{subfigure}
  \begin{subfigure}{\columnwidth}
      \centering
    \includegraphics[width=7.5cm]{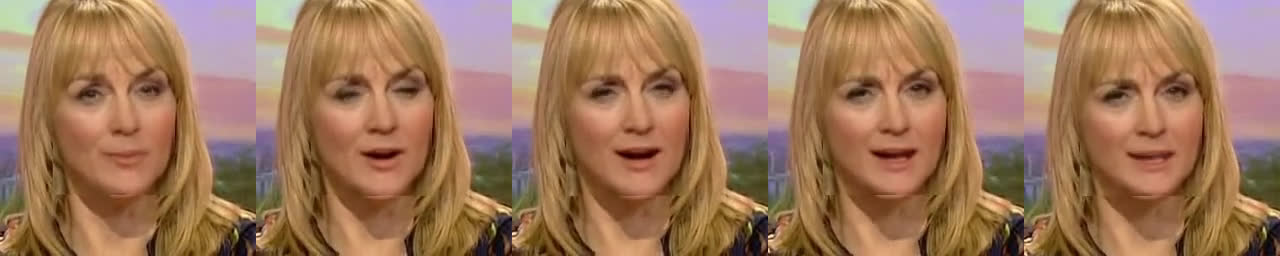}
    \caption{Faceformer}
  \end{subfigure}
  \begin{subfigure}{\columnwidth}
      \centering
    \includegraphics[width=7.5cm]{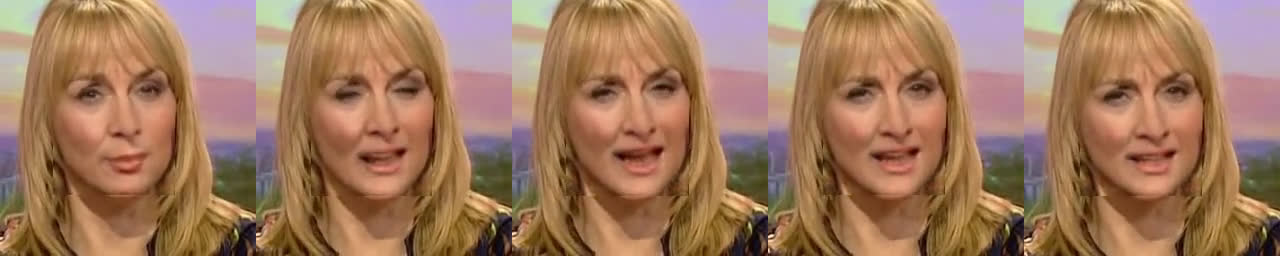}
    \caption{TalkLip ($g+c$)}
  \end{subfigure}
  \caption{Qualitative comparison.}
    \label{fig:distri_audio_ag1}
\end{figure}

\begin{figure}[!htb]
  \centering
  \begin{subfigure}{\columnwidth}
    \centering
    \includegraphics[width=7.5cm]{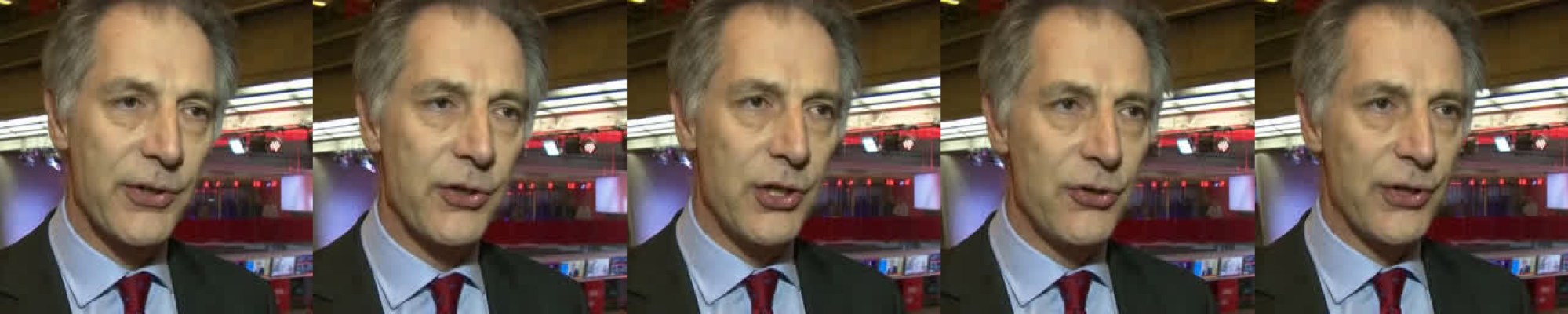}
    \caption{Ground Truth}
  \end{subfigure}
  \begin{subfigure}{\columnwidth}
      \centering
    \includegraphics[width=7.5cm]{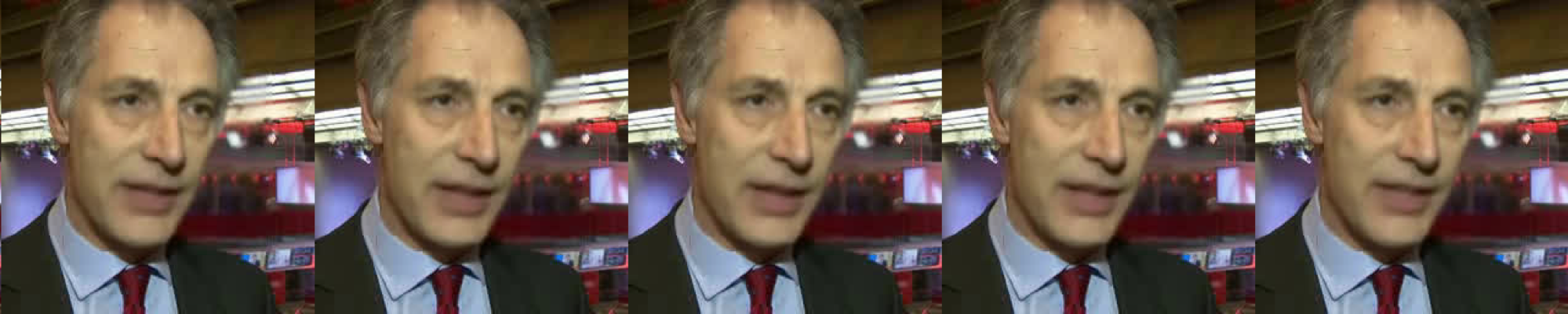}
    \caption{ATVGnet}
  \end{subfigure}
  \begin{subfigure}{\columnwidth}
      \centering
    \includegraphics[width=7.5cm]{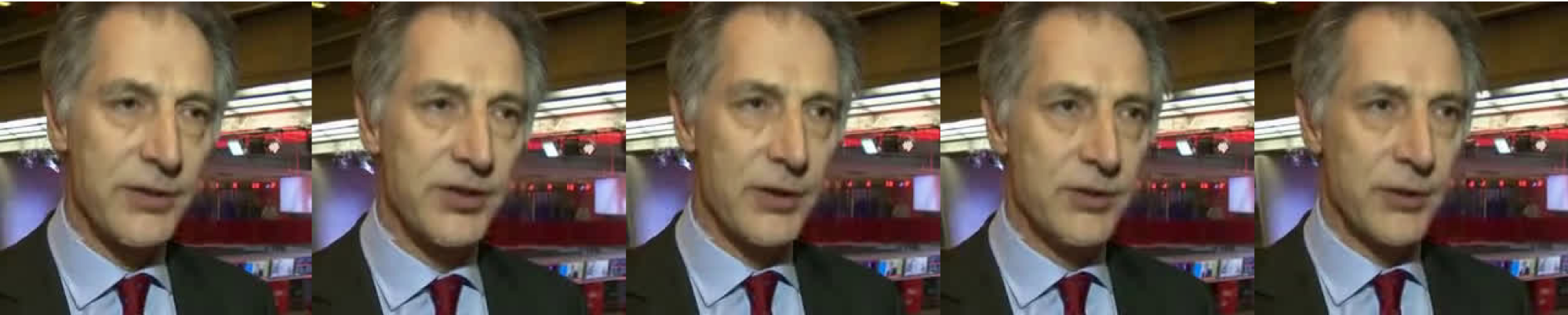}
    \caption{Wav2Lip)}
  \end{subfigure}
  \begin{subfigure}{\columnwidth}
      \centering
    \includegraphics[width=7.5cm]{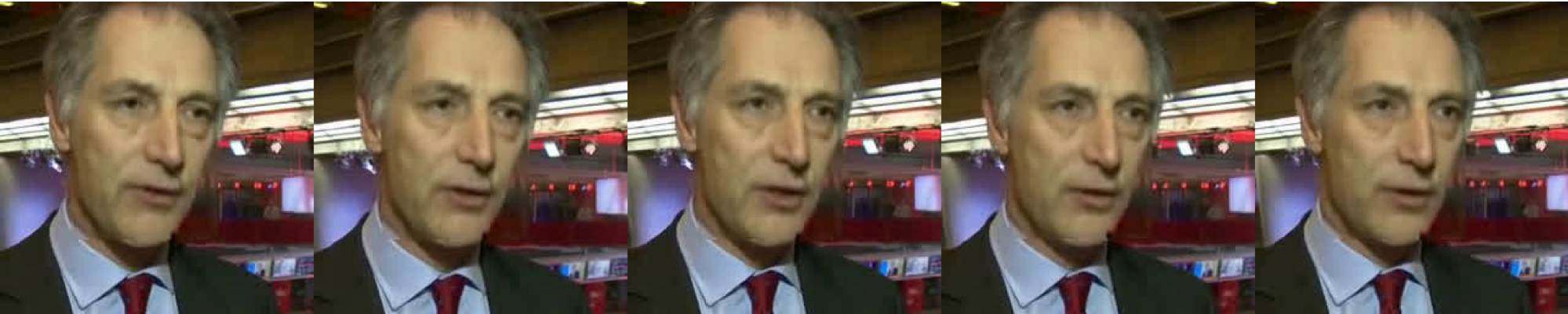}
    \caption{Faceformer}
  \end{subfigure}
  \begin{subfigure}{\columnwidth}
      \centering
    \includegraphics[width=7.5cm]{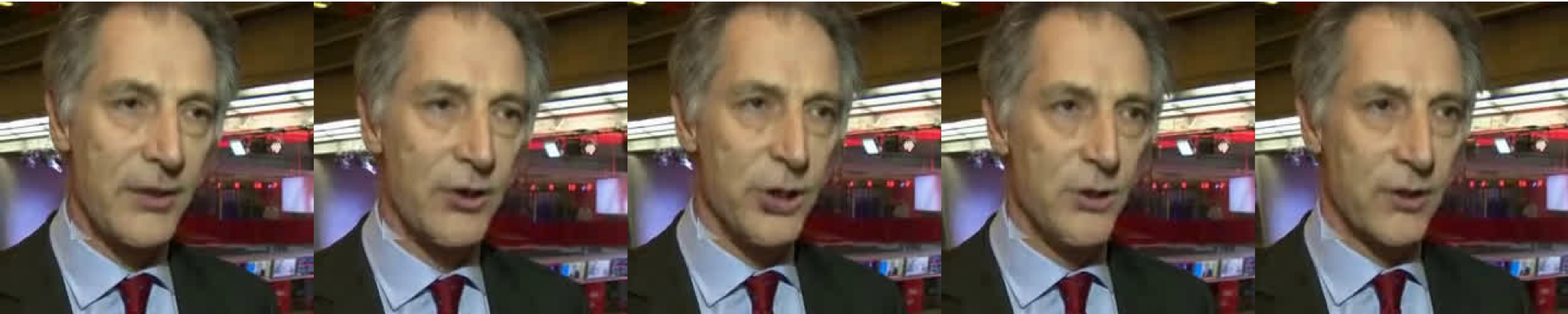}
    \caption{TalkLip ($g+c$)}
  \end{subfigure}
  \caption{Qualitative comparison.}
    \label{fig:distri_audio_ag2}

\end{figure}

\begin{figure}[!tb]
  \centering
  \begin{subfigure}{\columnwidth}
    \centering
    \includegraphics[width=7.5cm]{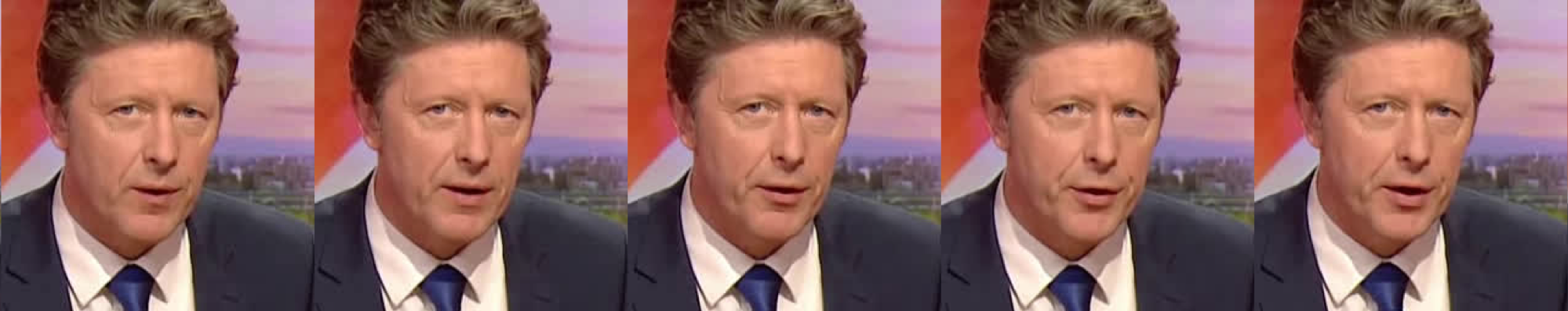}
    \caption{Ground Truth}
  \end{subfigure}
  \begin{subfigure}{\columnwidth}
      \centering
    \includegraphics[width=7.5cm]{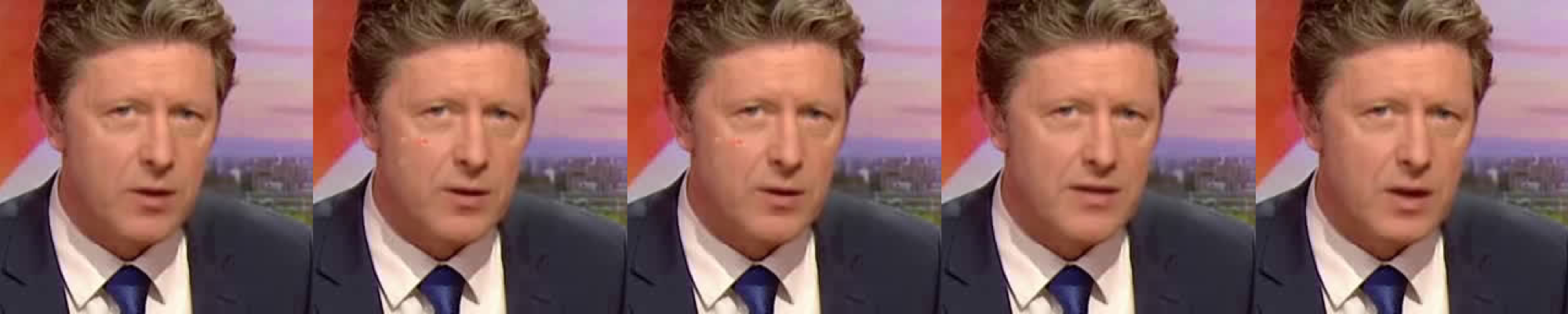}
    \caption{ATVGnet}
  \end{subfigure}
  \begin{subfigure}{\columnwidth}
      \centering
    \includegraphics[width=7.5cm]{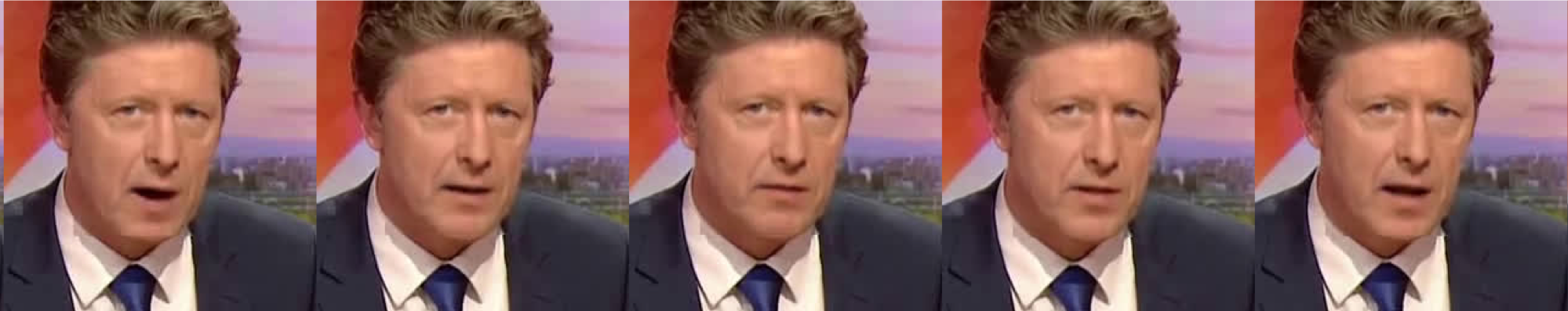}
    \caption{Wav2Lip)}
  \end{subfigure}
  \begin{subfigure}{\columnwidth}
      \centering
    \includegraphics[width=7.5cm]{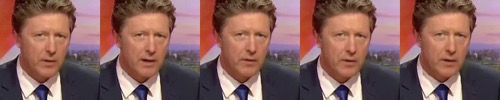}
    \caption{Faceformer}
  \end{subfigure}
  \begin{subfigure}{\columnwidth}
      \centering
    \includegraphics[width=7.5cm]{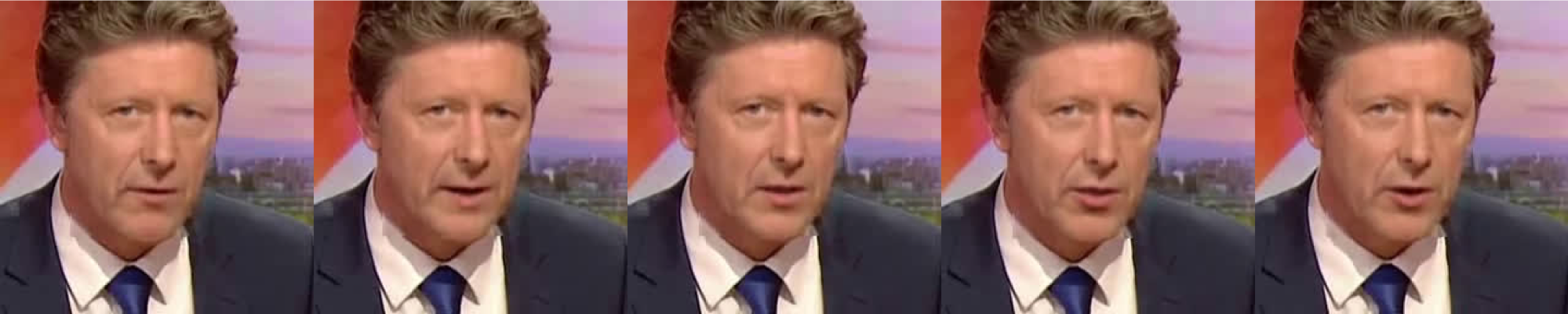}
    \caption{TalkLip ($g+c$)}
  \end{subfigure}
  
  \caption{Qualitative comparison.}
    \label{fig:distri_audio_ag3}

\end{figure}

\begin{figure}[!tb]
  \centering
  \begin{subfigure}{\columnwidth}
    \centering
    \includegraphics[width=7.5cm]{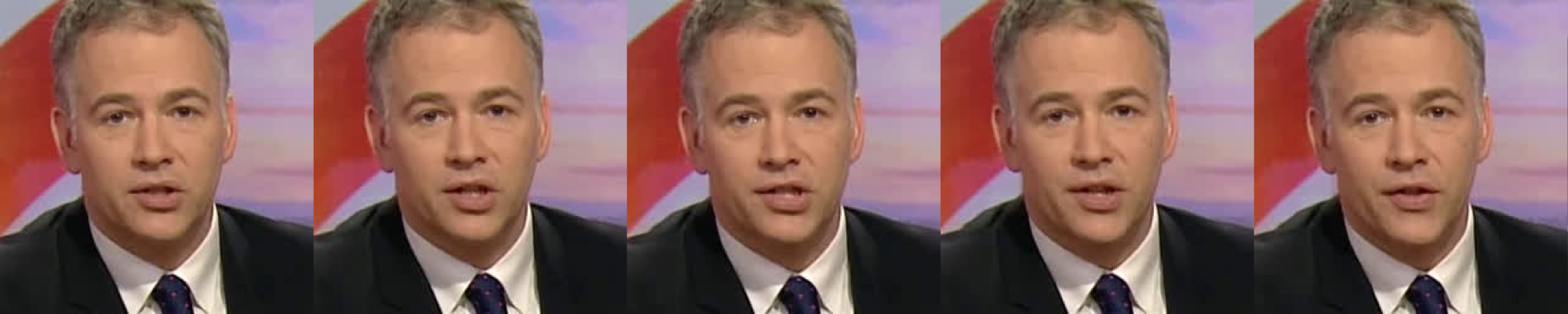}
    \caption{Ground Truth}
  \end{subfigure}
  \begin{subfigure}{\columnwidth}
      \centering
    \includegraphics[width=7.5cm]{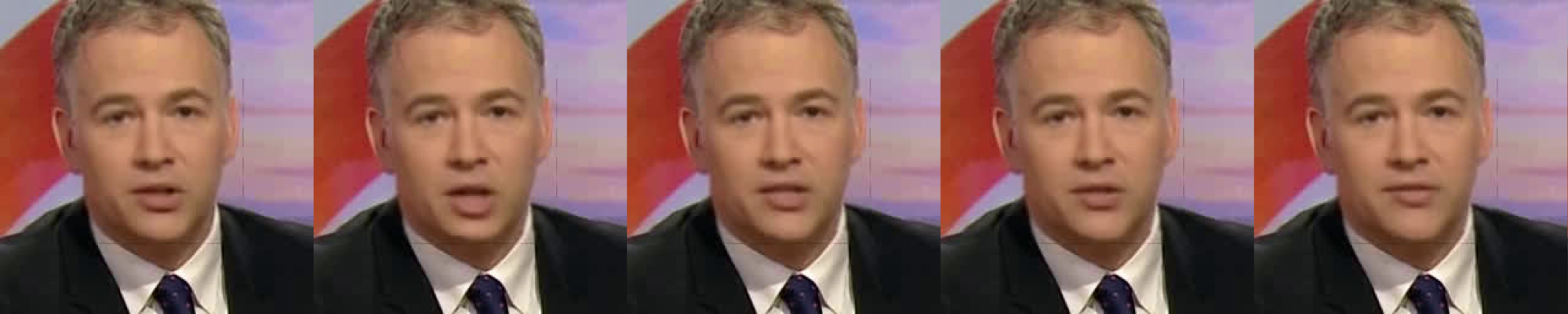}
    \caption{ATVGnet}
  \end{subfigure}
  \begin{subfigure}{\columnwidth}
      \centering
    \includegraphics[width=7.5cm]{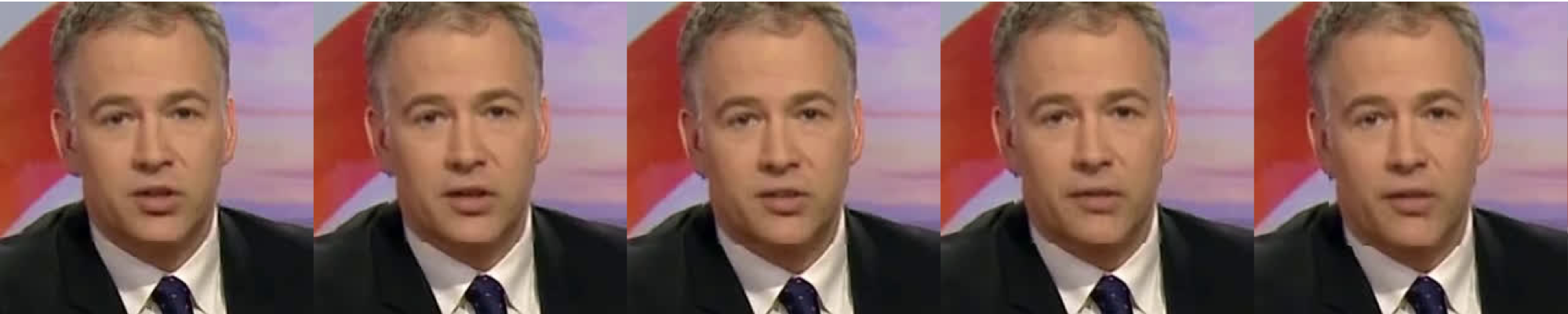}
    \caption{Wav2Lip)}
  \end{subfigure}
  \begin{subfigure}{\columnwidth}
      \centering
    \includegraphics[width=7.5cm]{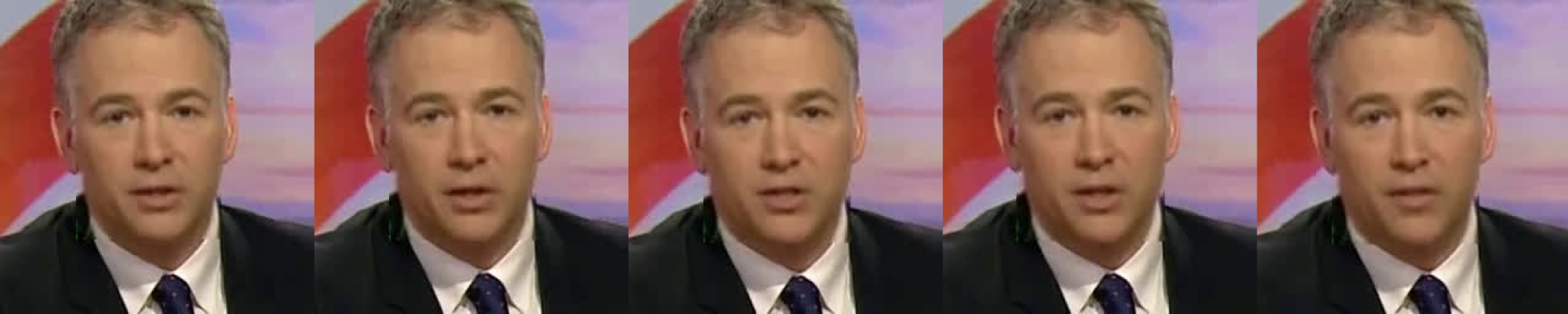}
    \caption{Faceformer}
  \end{subfigure}
  \begin{subfigure}{\columnwidth}
      \centering
    \includegraphics[width=7.5cm]{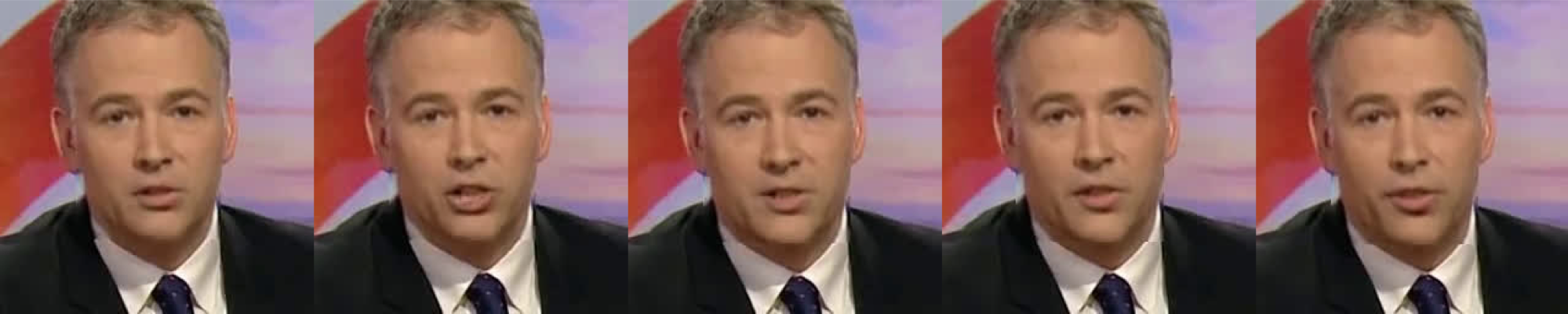}
    \caption{TalkLip ($g+c$)}
  \end{subfigure}
  
  \caption{Qualitative comparison.}
    \label{fig:distri_audio_ag4}

\end{figure}

In this section, we provide more qualitative comparisons with 3 State-of-Arts methods: ATVGnet\cite{chen2019sound}, Wav2Lip\cite{prajwal2020lip}, Faceformer\cite{fan2022faceformer} to show the superiority of our proposal. Please see details in Fig. \ref{fig:distri_audio_ag1}-\ref{fig:distri_audio_ag4}. 

\clearpage
\newpage

\vspace{-5.5 mm}

\vspace{+5.5 mm}

\end{document}